\documentclass[11pt, a4paper, logo, copyright, nonumbering]{baidusearch}
\usepackage[authoryear, sort&compress, round]{natbib}
\usepackage{dblfloatfix}
\usepackage[normalem]{ulem}
\usepackage{caption}
\usepackage{dramatist}
\usepackage{xspace}
\usepackage{pifont} 
\usepackage{tcolorbox}
\usepackage{xltabular}
\usepackage{longtable}
\usepackage{hyperref}
\usepackage{lineno}
\usepackage{multirow}
\usepackage{adjustbox}
\usepackage{fancyhdr}
\usepackage[bottom]{footmisc}

\usepackage{CJKutf8}
\usepackage{setspace}
\usepackage{wrapfig}

\usepackage{dsfont}
\usepackage{array} 
\usepackage{tabularx} 
\usepackage{subfigure} 
\usepackage{xcolor} 
\usepackage{booktabs}

\usepackage{graphicx}
\usepackage{amsmath,amssymb,amsfonts}
\usepackage{amssymb}
\usepackage{textcomp}
\usepackage{xcolor}
\usepackage[normalem]{ulem}
\usepackage{xfrac}
\usepackage{color}
\usepackage{tcolorbox}
\usepackage{booktabs}
\usepackage{makecell}
\usepackage{enumitem}
\usepackage{CJKutf8}
\usepackage{subfigure}
\usepackage{setspace}
\newcommand{\ie}{\textit{i.e.}}

\makeatletter
\newcommand{\thickhline}{%
    \noalign {\ifnum 0=`}\fi \hrule height 1pt
    \futurelet \reserved@a \@xhline
}
\newcolumntype{P}[1]{>{\centering\arraybackslash}p{#1}}

\def\@BTrule[#1]{%
  \ifx\longtable\undefined
    \let\@BTswitch\@BTnormal
  \else\ifx\hline\LT@hline
    \nobreak
    \let\@BTswitch\@BLTrule
  \else
     \let\@BTswitch\@BTnormal
  \fi\fi
  \global\@thisrulewidth=#1\relax
  \ifnum\@thisruleclass=\tw@\vskip\@aboverulesep\else
  \ifnum\@lastruleclass=\z@\vskip\@aboverulesep\else
  \ifnum\@lastruleclass=\@ne\vskip\doublerulesep\fi\fi\fi
  \@BTswitch}
\makeatother

\addto\extrasenglish{
}

 {\begin{list}{}%
         {\setlength{\leftmargin}{#1}}%
         \item[]%
 }
 {\end{list}}
 
\reportnumber{001} 

\title{\centering Towards AI Search Paradigm}

\author[*]{
Yuchen Li, Hengyi Cai, Rui Kong, Xinran Chen, Jiamin Chen, Jun Yang, Haojie Zhang, Jiayi Li, Jiayi Wu, Yiqun Chen, Changle Qu, Wenwen Ye, Lixin Su, Xinyu Ma, Lingyong Yan, Long Xia, Daiting Shi, Junfeng Wang, Xiangyu Zhao, Jiashu Zhao, Haoyi Xiong, Shuaiqiang Wang,\newline Dawei Yin$^{\dag}$
\\
\normalsize
Baidu Search
\\
\normalsize
\texttt{\{yuchenli13, yindawei\}@acm.org}
}
\correspondingauthor{\small $^{\dag}$Corresponding author}

\begin{abstract}

 In this paper, we introduce the AI Search Paradigm, a comprehensive blueprint for next-generation search systems capable of emulating human information processing and decision-making. The paradigm employs a modular architecture of four LLM-powered agents (Master, Planner, Executor and Writer) that dynamically adapt to the full spectrum of information needs, from simple factual queries to complex multi-stage reasoning tasks. These agents collaborate dynamically through coordinated workflows to evaluate query complexity, decompose problems into executable plans, and orchestrate tool usage, task execution, and content synthesis. We systematically present key methodologies for realizing this paradigm, including task planning and tool integration, execution strategies, aligned and robust retrieval-augmented generation, and efficient LLM inference, spanning both algorithmic techniques and infrastructure-level optimizations. By providing an in-depth guide to these foundational components, this work aims to inform the development of trustworthy, adaptive, and scalable AI search systems.

\end{abstract}

\begin{document}

\begin{CJK*}{UTF8}{gbsn}

\maketitle

\vspace{2ex}
{\keywordsfont\noindent\textbf{Keywords:} AI Search, Web Search, Agentic Search, Search Engine, Large Language Models, Retrieval-Augmented Generation, Task Planning, Multi-Agent System}

\newpage

\begin{spacing}{0.9}
\tableofcontents
\end{spacing}

\newpage
\section{Introduction}

In an age where individuals are constantly managing an overwhelming flow of data, \emph{Information Seeking} (IS), an active behavior that constructs new cognition when faced with knowledge gaps \citep{wilson1981user, dervin1983overview}, has emerged as a vital process for making informed decisions and solving complex problems. The advent of web search engines marked a significant leap in IS. Web search engines function as specialized systems built upon fundamental \emph{Information Retrieval} (IR) technologies, systematically crawling, indexing, and retrieving internet-based information in response to user queries.
Over the past decades, the field of IR has undergone transformative generational shifts to achieve superior performance. Researchers and practitioners have progressed systematically from early lexical models to machine learning paradigms, and ultimately, to Large Language Models (LLMs).

Lexical IR technologies predominantly rely on keyword matching techniques, encompassing vector space models, probabilistic frameworks, and traditional language models \citep{brin1998anatomy, DBLP:journals/arist/Yang05, robertson2009probabilistic, mikolov2013efficient}. These approaches represent documents and queries as bags of words, estimating relevance based on the exact or partial overlap of terms. While effective for exact term matching, they inherently struggle with semantic mismatches and vocabulary variation—failing, for example, to associate synonymous expressions or contextual nuances. Such limitations have increasingly highlighted the need for more robust retrieval mechanisms. This motivated the field's transition toward semantic-aware retrieval paradigms capable of deeper information assimilation.

Learning-to-Rank (LTR) methodologies \citep{RankNet:pairwise, ListNet:listwise, DBLP:journals/ir/QinLL10} emerged from the pursuit of more relevant and higher-quality search results, driving the application of machine learning to ranking problems. 
This next generation of information retrieval systems replaced heuristic ranking approaches with machine-learned models that directly optimize ranking objectives. LTR techniques leveraged feature engineering---incorporating signals including text-matching scores, document structure metadata, authority metrics, and user behavior signals (e.g., click-through rates)---combined with supervised machine learning to optimize relevance ranking. This represented a significant advancement in ranking effectiveness, consistently placing more relevant documents in top positions compared to lexical retrieval models. The output of such systems is a ranked list of documents, requiring users to perform an additional step to access the target information: clicking on relevant documents and then locating or synthesizing the needed content --- revealing a gap between system outputs and true information needs.

The advent of LLMs \citep{wei2022chain, achiam2023gpt, bubeck2023sparks, DBLP:journals/corr/abs-2403-08295, DBLP:journals/corr/abs-2407-21783} has enabled information retrieval systems to bridge this gap, transitioning from document retrieval to direct generation of precise contextualized answers. 
However, current RAG systems primarily function as single-shot answer generators and struggle with queries involving complex information needs, such as innovation-related inquiries, emotional nuance, or proactive planning, the orchestration of diverse tools, and deep reasoning across multiple, sometimes conflicting, pieces of evidence \citep{DBLP:conf/kdd/FanDNWLYCL24, DBLP:journals/corr/abs-2402-19473}. 
These systems' output quality is critically dependent on the initial document retrieval phase, lacking both robustness in handling imperfect retrieval and the sophisticated reasoning capabilities needed to address truly multifaceted user needs. 
For instance, a seemingly straightforward question like "Who was older, Emperor Wu of Han or Julius Caesar, and by how many years?" poses significant challenges for current systems. While individual birth dates for both historical figures may be easily retrievable, no single document explicitly compares their ages. This requires the system to perform a complex, multi-stage process: (1) retrieving and verifying each emperor's birth year from separate sources, (2) resolving potential conflicts between contradictory records, (3) calculating the age difference, and (4) synthesizing the final comparison. Such queries demand multi-step reasoning capabilities—the ability to decompose complex questions, execute sequential sub-queries, evaluate evidence reliability, and integrate intermediate results into a coherent answer.

This paper establishes that the historical evolution of IR systems necessitates a fundamental advancement: cognitive architectures that genuinely emulate human information foraging behaviors and multi-stage reasoning processes. To meet this need, we propose a revolutionary IS paradigm, \textbf{AI Search Paradigm} — a collaborative, multi-agent framework powered by large language models that reasons, plans, and executes complex problem — solving strategies on the user’s behalf. Drawing inspiration from early work on computational algorithms \citep{Knuth97} and system interactions \citep{Cohen07}, this AI search paradigm coordinates specialized agents, each responsible for a distinct phase of the information-seeking process, to jointly deliver accurate, context-rich answers that mirror human-level inquiry. The key agent roles include:

%

\begin{itemize}

    \item \textbf{Master Agent.}   \emph{Master} serves as the initial point of contact, responsible for analyzing the user’s query to assess its complexity and intent. Based on the nature and difficulty of the problem, \emph{Master} dynamically coordinates and assembles an appropriate team of subsequent agents. This is a unique feature of the proposed paradigm, compared to traditional IR systems, which incorporate static query understanding modules~\citep{xiong2024search}, and RAG systems, where initial query analysis typically leads to a fixed processing pipeline~\citep{ma2023query}. In addition, it continuously evaluates the performance of subordinate agents; in the event of task failures, \emph{Master} conducts reflective analysis and guides the team to re-plan and re-execute accordingly.



    \item \textbf{Planner Agent.} \emph{Planner} is only invoked for complex queries that require multi-step reasoning or information gathering.
   Given the user's input query, it selects appropriate tools from tool platforms—e.g., the conceptualized Model-Context Protocol (MCP) Servers Platform~\citep{edwards2025mcp}—to dynamically adjust the capability boundaries of the LLMs. \emph{Planner} then decomposes the overarching query into a structured sequence of manageable sub-tasks, represented as a Directed Acyclic Graph (DAG). This step involves strategic planning of how to approach the problem. While advanced RAG systems might perform query decomposition or multi-hop reasoning~\citep{ma2023query}, \emph{Planner}'s function of generating a comprehensive, explicit DAG of sub-tasks with interdependencies and potential tool bindings represents a more flexible and proactive planning capability than typically found in standard RAG systems~\citep{DBLP:conf/nips/LewisPPPKGKLYR020}.

    \item \textbf{Executor Agent.} \emph{Executor} is responsible for carrying out the simple query or the individual sub-tasks defined by \emph{Planner}. A crucial aspect of its role is the invocation of external tools, selected by \emph{Planner} from tool platforms (e.g. MCP), to gather information or perform computations necessary for each sub-task. It also evaluates the outcomes of these executions. Traditional IR systems execute internal algorithms, and while some RAG models can use tools, \emph{Executor}'s role in managing a diverse set of tools for DAG-derived sub-tasks, including outcome assessment and fallback mechanisms, suggests a more robust and orchestrated execution layer~\citep{quexploration}.

    \item \textbf{Writer Agent.} The final agent in the workflow, \emph{Writer}, synthesizes the information and results gathered from all completed sub-tasks. Its goal is to generate a coherent, contextually rich, and potentially multi-perspective response to the user's original query, including necessary filtering and disambiguation. While the generator component in RAG systems also synthesizes answers~\citep{DBLP:conf/nips/LewisPPPKGKLYR020}, \emph{Writer} in the AI search paradigm is envisioned to work with a more structured and diverse set of inputs from various planned sub-tasks, aiming for a more comprehensive and conversational synthesis than generation from a flat list of retrieved documents.

\end{itemize}
This paradigm emphasizes adaptability through dynamic team configurations and aims to achieve comprehensive and context-aware information synthesis, moving beyond simple document retrieval \citep{Abril07}.

This paper presents a comprehensive exploration of the proposed \textbf{AI Search paradigm}. Rather than focusing on a singular system and its evaluation, the contributions of this work
lie in the exploration of essential techniques for the AI Search paradigm and can be summarized as follows:

\begin{itemize}

    \item \textbf{Conceptualizing a New Paradigm.} 
    We introduce and detail a dynamic, modular multi-agent architecture for AI Search, featuring specialized agents that perform distinct search-oriented roles (e.g. mastering, planning, execution, and synthesis). The system dynamically coordinates these agents on the fly to adapt to query complexity, enabling more flexible and powerful reasoning capabilities than traditional approaches. This represents a significant departure from both the monolithic models of conventional web search and the ``linear'' pipelines of Retrieval-Augmented Generation (RAG) systems.


    \item \textbf{Foraging Core Agentic Search Methodologies.} We identify and contextualize critical techniques for the core AI search process. This includes methods for \emph{Advanced Task Planning}, including dynamic adjustment of LLMs' capability boundary, advanced tool retrieval methods, DAG-based task planning, \emph{Master}-guided reflection and re-action mechanism, and reinforcement learning-enhanced optimization strategy. It also covers \emph{Flexible Task Execution strategies} that focus on aligning evidence retrieval with LLM's preferences, lightweighting retrieval and ranking systems, and constructing LLM-augmented features.
    
    \item \textbf{Detailing Search-Oriented Generation and Optimization Techniques.} We subsequently examine methodologies for generating robust responses and optimizing the overall system. This discussion encompasses strategies for ensuring both robustness and alignment, as well as a joint multi-module optimization approach tailored for RAG systems. Finally, we address \emph{Efficient LLM Inference Strategies}, termed ``Lightning LLM's Generation'', which include optimizations at both the algorithmic and infrastructure levels.

    \item \textbf{Demonstrating Improved User Experience Through Case Studies.} To showcase the practical advantages of our paradigm, this paper conducts extensive experiments and presents several case studies evaluated through user experience studies. Instead of relying on statistics of performance metrics, these cases illustrate how the system handles complex, multi-step queries from a user's perspective. By walking through these examples, we demonstrate the qualitative improvements in the search experience, highlighting the system's ability to deliver more comprehensive, accurate, and satisfying answers compared to previous technologies.

\end{itemize}
The overarching contribution of this paper is thus to provide a structured and detailed blueprint that assembles best-of-breed techniques from industry and academia, outlining an evolving design paradigm that can serve as a guide for future research and development in AI-driven search.

\begin{figure*}[!t]
    \centering
    \includegraphics[width=\linewidth]{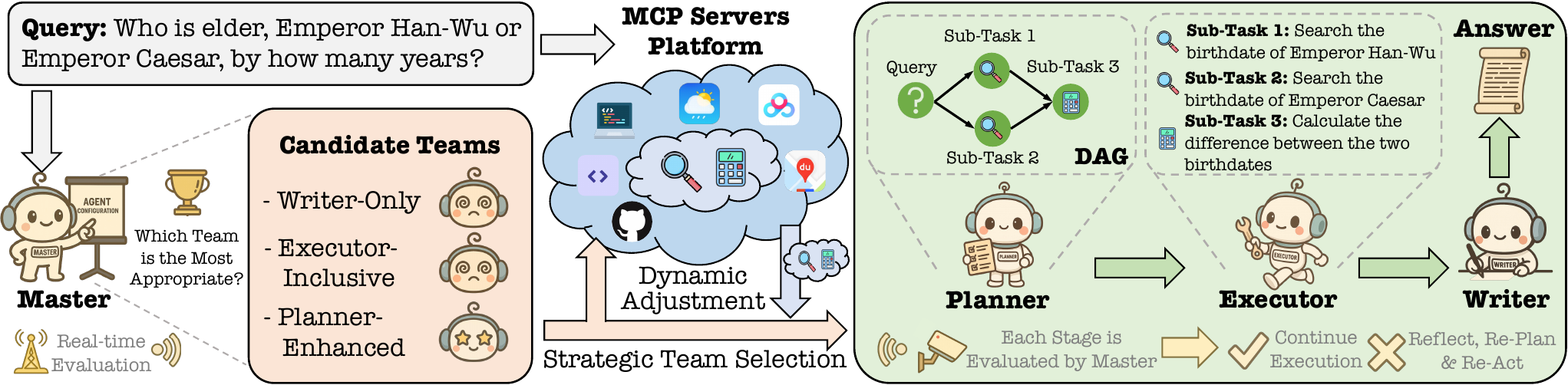}
    \caption{\centering The Overview of AI Search Paradigm.}
    \label{fig:overall_framework_1}
\end{figure*}

\section{System Overview}\label{sec:overview}

In this section, we introduce a modular AI search paradigm designed to address the limitations of conventional retrieval-augmented systems in handling complex, multi-step information needs. Inspired by human collaborative search behavior, the framework coordinates multiple specialized agents to evaluate, plan, execute, and synthesize in a scalable and context-aware manner. Specifically, the agents are assigned a range of specialized responsibilities, including proactively evaluating the complexity of user queries, configuring appropriate multi-agent teams, conducting task decomposition and planning, selecting suitable tools via tool platforms (e.g. the MCP servers platform), executing the selected tools to fulfill specific sub-tasks, and synthesizing the outcomes of all sub-tasks to generate comprehensive and multi-perspective responses. Such a coordinated architecture ensures a methodical progression of the search process toward coherent and comprehensive outcomes.

The multi-agent setting offers significant advantages over single-agent configurations, which often suffer from task overload and diminished efficiency when a single LLM-based agent is responsible for managing multiple complex responsibilities. By assigning well-defined roles to each agent, the system ensures clear task allocation and robust operational management, thereby preventing bottlenecks and enhancing overall performance. Moreover, dynamically configuring agent teams in response to varying levels of query complexity enhances the system’s flexibility and adaptability across diverse information-seeking tasks. Building on this, the modular design further reinforces scalability, enabling the system to maintain consistently high performance in a wide range of complex and demanding scenarios.

\textbf{Details of Agent Roles.} 
As illustrated in Fig.~\ref{fig:overall_framework_1}, the AI search paradigm is implemented as a multi-agent collaborative system. 
Based on their designated functional roles, the system comprises four specialized agents: \emph{Master}, \emph{Planner}, \emph{Executor}, and \emph{Writer}. The specific tasks assigned to each agent are detailed as follows:
\begin{itemize}
    \item \textbf{\emph{Master.}} 
    In the AI Search system, the Master agent acts as the team coordinator. It is responsible for analyzing the user's input query to assess its complexity and intent, and for assembling an appropriate team of specialized agents to handle the task --- without directly participating in the downstream query processing itself. Specifically, the Master agent analyzes the input query along with relevant contextual information to assess its complexity and determine the optimal team configuration. For relatively simple queries, it may assign a minimal setup involving only the Writer or a combination of the Executor and Writer. For more complex queries, it initiates a larger team that includes the Planner agent to perform task decomposition and planning, enabling a step-by-step resolution of the query. Furthermore, the Master continuously monitors the performance of subordinate agents. In the event of task failure, it conducts a reflective analysis and directs the team to re-plan and re-execute accordingly.
    \item \textbf{\emph{Planner.}} 
    This agent is only invoked for complex queries that require multi-step reasoning or information gathering, and is responsible for task decomposition and strategic planning in response to complex queries, enabling the AI search system to address them through a sequence of simpler, structured sub-tasks. Its core function is to select appropriate tools from a tool set (e.g. the MCP platform) to dynamically adjust the LLMs' capability boundaries and to generate a DAG that encodes a global task plan. In this DAG, each node represents an atomic, schedulable sub-task, and each edge captures a conditional dependency between sub-tasks, derived from the Planner’s reasoning process.
    Furthermore, when the Master evaluates sub-tasks as having execution errors or lacking critical data, the Planner, under the Master's direction, undertakes a reconfiguration of the DAG.
    \item \textbf{\emph{Executor.}} In the task execution phase, the Executor plays a central role in carrying out the task and evaluating its outcome, ensuring successful completion and compliance with predefined requirements. Each sub-task arrives at execution with either an external tool pre-assigned or a ``tool-free'' configuration to be handled solely by the Executor using its own built-in capabilities. During execution, the Executor may repeatedly invoke the assigned tool based on task complexity, while continuously assessing whether the accumulated outputs are sufficient to fulfill the sub-task’s objectives. Once the evaluation indicates that the requirements are satisfied, the current output is returned; otherwise, the execution continues iteratively. In cases where the assigned tool becomes unresponsive or fails, the system seamlessly switches to a functionally equivalent backup tool within the same tool module, thereby maintaining execution continuity and robustness. For example, consider a sub-task that requires acquiring relevant knowledge via a web search tool. The Executor invokes the bound \emph{web search tool} using the generated sub-queries based on the current sub-task and performs multiple rounds of search as needed. After each round, it evaluates the retrieved documents to determine whether the information sufficiently supports the factual or contextual requirements of this sub-task. If not, these sub-queries are refined and the tool is re-invoked. Once adequate coverage is achieved, the results are finalized and returned. Through this integrated approach, the Executor effectively combines execution and quality control within a unified mechanism, ensuring robust, accurate, and context-aware execution of sub-tasks.
    \item \textbf{\emph{Writer.}} This agent is responsible for synthesizing the information and results gathered from all completed preceding sub-tasks and generating the final response to the user's query. Its goal is to generate a coherent, contextually rich, and potentially multi-perspective response to the user's original query, including necessary filtering and disambiguation. During this process, it identifies the semantic relationships and logical structures among the outputs of multiple sub-tasks, and reorganizes the information using either predefined templates or adaptive structuring strategies. This enables the Writer to produce a response that is coherent, well-organized, and easy to comprehend. When redundancy or semantic inconsistency exists across sub-task results, the Writer performs content filtering and disambiguation to ensure that the final output is accurate and free from misleading information. Beyond simply answering the core query, the generated response also incorporates contextual background and explanatory content when necessary, thereby improving the completeness of the answer and enhancing overall user satisfaction.
\end{itemize}

\textbf{Workflow of Multi-Agents.}
    According to the complexity of the query and the specific functions of these agents, the AI search system adopts three distinct team configurations to support different levels of reasoning and execution: (1) \emph{Writer-Only Configuration}, (2) \emph{Executor-Inclusive Configuration}, and (3) \emph{Planner-Enhanced Configuration}. The operational details of each configuration are described as follows:
\begin{itemize}
    \item \textbf{Writer-Only Configuration.} This configuration is suitable for the simple query that can be answered directly based on the system (LLM)’s reasoning capability and internalized knowledge. In this case, the Master analyzes the query and determines that neither external tool invocation nor task decomposition is necessary. It then directly delegates the task to the Writer, who generates a response relying solely on its built-in generative capabilities. For example, given the query ``\emph{What is the name of Emperor Han-Wu?}'', the Master forwards the prompt to the Writer, who directly generates the response ``\emph{The name of Emperor Han-Wu is Liu Che.}''
    \item \textbf{Executor-Inclusive Configuration.} This configuration handles moderately complex queries that require factual information from external sources but do not involve multi-step reasoning or decomposition. The Master determines that the query can be resolved through single-step execution and assigns it directly to the Executor. Meanwhile, the AI search system uses the query semantics along with contextual cues derived from the team configuration to query the tool platform server (e.g. MCP server), retrieving a focused and semantically relevant subset of tools. The Executor selects the top-ranked tool from this subset and invokes it to perform the task. Upon obtaining the result, it evaluates whether the output meets the task requirements. Once execution is complete, the output is forwarded to the Writer, who synthesizes the final response. 
    For instance, given the query ``\emph{Is Beijing's weather good for going out today?}'', the Executor invokes the retrieved \emph{weather query tool}, obtains real-time weather data, and after verifying its completeness, passes the result to the Writer. The response might be: ``\emph{Beijing’s weather today is sunny, with temperatures ranging from 12°C to 25°C. It is suitable for outdoor activities, although precautions against UV exposure are advised.}''
    \item \textbf{Planner-Enhanced Configuration.} This configuration is designed for complex queries that require multi-step reasoning and structured task execution. When high query complexity is detected, the Master delegates control to the Planner, initiating a planning-based workflow. In parallel, tool selection by the Planner is supported by querying the tool platform server (e.g. MCP server) for a semantically filtered candidate set, based on the input query and current execution context. The Planner decomposes the query into a sequence of atomic sub-tasks and organizes them into a DAG that captures their logical and execution dependencies. For each sub-task, a suitable tool is selected from the retrieved tool subset and explicitly bound to the corresponding node. The Executor then traverses the DAG layer by layer, invoking the bound tools and evaluating the intermediate results. Once all sub-tasks are completed, the aggregated outputs are passed to the Writer, who synthesizes them into a coherent and context-aware final response.
    For example, consider a complex query ``\emph{Who is elder, Emperor Han-Wu or Emperor Caesar, by how many years?}''. In this scenario, the Master delegates the query to the Planner, who is responsible for both planning and decomposing it into three specific sub-tasks, each associated with a specific tool from the candidate set: 
    \begin{itemize}
        \item \emph{Sub-Task 1: Search the birthdate of Emperor Han-Wu. Using the Web Search Tool.}
        \item \emph{Sub-Task 2: Search the birthdate of Emperor Caesar. Using the Web Search Tool.}
        \item \emph{Sub-Task 3: Calculate the difference between the two birthdates. Using the Programmer Tool.}
    \end{itemize} 
    These sub-tasks are structured into a DAG according to their execution dependencies and are executed sequentially by the Executor. Once all results are collected, the Writer synthesizes them into a coherent and contextually accurate response: ``\emph{Emperor Wu of Han (156–87 BC) lived for approximately 69 years, whereas Julius Caesar (100–44 BC) lived for about 56 years. Emperor Wu of Han was therefore older by approximately 56 years.}''
\end{itemize}
Overall, these configurations illustrate the system’s adaptive execution flow, where different agent teams are engaged based on query complexity. By selecting suitable configurations that range from direct generation for simple tasks to tool-assisted multi-step planning for complex queries, the system ensures both efficiency and scalability across diverse search scenarios.

In the remainder sections, we provide a detailed exploration of the implementation algorithms and optimization strategies applied to these agents.

\section{Task Planner}
\begin{figure*}[t]
    \centering
    \includegraphics[width=0.95\textwidth]{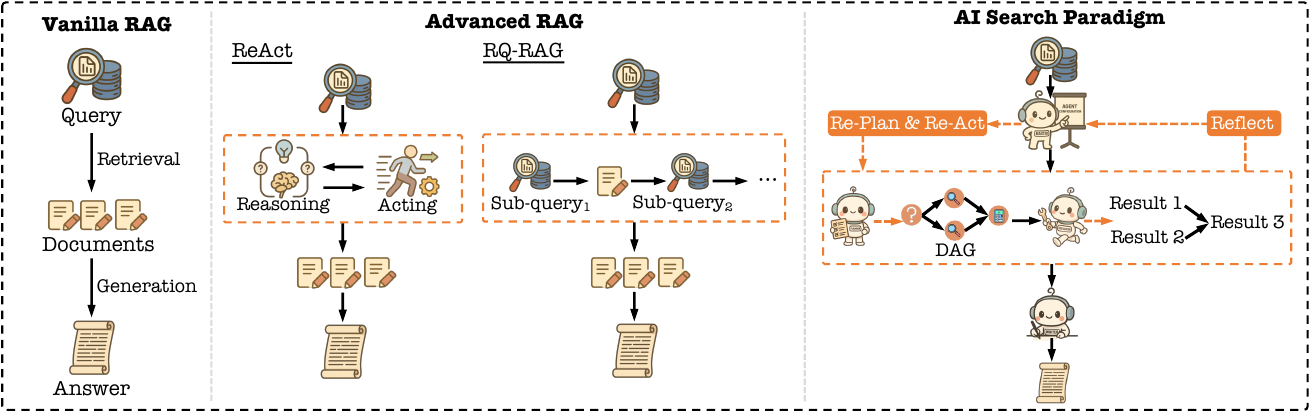}
    \caption{Comparison of RAG frameworks. \textbf{Left:} Vanilla RAG conducts one-shot retrieval followed by direct answer generation. \textbf{Middle:} Advanced RAG methods, such as ReAct and RQ-RAG, involve reasoning-action cycles or sequential sub-query execution. \textbf{Right:} AI search paradigm introduces a multi-agent system wherein \emph{Master} guides \emph{Planner} to formulate a plan based on the input query, while also continuously evaluating the execution status and completeness of sub-task results, and performing reflection and re-planning when necessary. \emph{Planner} is responsible for constructing a DAG of sub-tasks and dynamically selecting the appropriate tools, thereby enabling structured and adaptive multi-step execution. \emph{Executor} executes the specific sub-tasks using these tools, and finally, \emph{Writer} generates the final answer.}
    \label{fig:comparison}
\end{figure*}
As the core reasoning component in the AI search system, the Planner is responsible for decomposing complex queries into structured sub-tasks and orchestrating execution through appropriate tools. Unlike conventional systems that rely on static retrieval and fixed response generation, the Planner enables dynamic task planning, effective management of multiple tools, and adaptive decision-making. In this section, we elaborate on its motivation, functional design, and optimization strategies within the broader context of AI Search paradigm.

\subsection{Background}
When addressing complex queries that necessitate multi-step sub-task execution and the coordinated invocation of multiple tools, traditional RAG systems often fall short in effective task planning and dynamic multi-tool coordination. Consequently, their reliance on single-step task execution or fixed-tool utilization frequently results in incomplete information retrieval or inaccurate responses. We revisit the motivating example from a large-scale real-world search engine in Section~\ref{sec:overview} to underscore the limitations of traditional systems and motivate the design of the Planner: ``\emph{Who is elder, Emperor Han-Wu or Emperor Caesar, by how many years?}''. To resolve this query, it is first decomposed into three specific sub-tasks, each assigned to a dedicated tool. Then, these sub-tasks are organized into a DAG that reflects their logical execution order and subsequently executed in sequence. Throughout this section, we will employ this example to provide a comparative analysis of how recent RAG systems and our proposed search system approach the same problem.

\textbf{Recent RAG Workflow.} 
A conventional RAG approach begins by submitting the query to a retriever to obtain relevant documents, as illustrated in the left panel of Fig.~\ref{fig:comparison}.
These documents, along with the query, serve as a prompt for an LLM to generate an answer. In practice, this method has yielded incorrect and incomplete results (\ie, returning “Emperor Caesar” in our case), due to incomplete retrieval and the failure to capture the necessary birth dates for Emperor Han-Wu and Emperor Caesar. This shortcoming primarily arises from a retrieval process that lacks prior reasoning, as the query only mentions the pertinent names without specifying the need for birth dates, leading the language model to generate inferences without appropriate factual support.

Moreover, advanced RAG approaches, as illustrated in the middle part of Fig.~\ref{fig:comparison} (\ie, ReAct~\citep{yao2023react}), introduce a structured loop that alternates between \emph{Thought}, \emph{Action}, and \emph{Observation} phases. In such frameworks, the model plans the subsequent reasoning step, retrieves relevant information, and iteratively incorporates the results. When applied to the current case, a ReAct-like agent would first consider retrieving the birth dates, then carry out document retrievals for both emperors, and update its internal state accordingly.  However, despite these enhancements, ReAct’s reliance solely on in-context memory precludes genuine external tool invocation; consequently, even if the agent correctly determines which emperor is older, it cannot precisely compute the age difference due to the absence of a dedicated calculator tool and structured numerical reasoning. Furthermore, the iterative context expansion inherent in ReAct complicates the maintenance of entity alignment and consistent task planning, thereby increasing the risk of error propagation.
Similarly, RQ-RAG~\citep{chan2024rq} seeks to enhance RAG performance by explicitly decomposing complex queries into sequential subqueries, solving them step-by-step while retaining intermediate results. For instance, a query decomposition-based agent would first retrieve Emperor Han-Wu’s birth date, then that of Emperor Caesar, and subsequently deduce their age difference. Nonetheless, despite the improved decomposition capabilities, RQ-RAG still operates within a single retrieval-and-generation framework that lacks dynamic tool utilization. As a result, although it may successfully obtain individual birth dates, it is unable to perform the subsequent computational reasoning necessary to determine the precise age difference, leaving the answer incomplete.
In summary, while ReAct and RQ-RAG represent significant advancements over traditional RAG architectures by incorporating iterative reasoning and query decomposition, their inability to dynamically invoke external tools and to plan tasks restricts their effectiveness in accurately resolving complex, multi-step queries.

\textbf{Necessity of Task Planner.}
To overcome these limitations, a dedicated Planner is indispensable for complex query resolution. Unlike previous RAG systems that rely exclusively on retrieval or in-context decomposition, a Planner explicitly decomposes queries into fine-grained sub-tasks, determines their logical dependencies via a DAG, and dynamically selects appropriate tools for task execution beyond simple retrieval, as illustrated in the right part of Fig.~\ref{fig:comparison}. Moreover, the planner is capable of re-planning; if any intermediate results deviate from the expected objectives, the Planner, under the guidance of the Master, adjusts the task plan accordingly. These capabilities fundamentally expand the scope of retrieval-augmented systems, transforming passive ``\emph{retrieve‐then-generate}'' pipelines into ``\emph{reason, plan, execute and re-plan}'', a proactive AI search system. 
Concretely, the planner must decide \underline{what} can be planned, \underline{which} tools should be invoked, and \underline{how} to optimize its own behaviour over time.


\subsection{Task Universe and the MCP Abstraction}
\label{ssec:mcp}
Early tool-augmented LLM systems relied on vendor-specific ``function-calling'' JSON schemas such as those introduced by OpenAI~\citep{openai2025responses,openaifunctioncalling}
and rapidly copied by many frameworks~\citep{langchainagentprotocol}.  
While simple, these ad-hoc contracts were \emph{coupled} to one provider, lacked machine-typed guarantees, and made it impossible for independent agents to share tools or reason about cost, latency, or security across organizational boundaries. For AI Search whose planner must orchestrate heterogeneous knowledge look-ups, computations, and transformations in a single multi-step plan, such fragmentation is a major bottleneck.

\textbf{Model-Context Protocol.} MCP addresses this fragmentation by specifying a vendor-neutral, HTTP+JSON-RPC interface through which \emph{Servers} expose tools and data, while \emph{Clients} (i.e., LLMs or agents) discover, invoke, and monitor those tools in a secure, typed fashion~\citep{anthropic2024mcp,schmid2025overview}.  
Specifically, the protocol bundles four ingredients: 
(1) a \textit{manifest} that advertises each endpoint’s name, semantic role, cost, and latency bounds; 
(2) machine-readable \textit{input/output schemas} that ground the LLM’s function-calling tokens; 
(3) a \textit{capability handshake} for tool discovery, and (4) an \textit{execution contract} that guarantees idempotent, auditable calls.


\subsection{Dynamic Capability Boundary}\label{ssec:what_planner}
\begin{minipage}{\textwidth}
Given a user's input query and MCP servers, we define the combination of the LLM for the Planner and the tool set as \emph{Capability Boundary}. Specifically, the capability boundary encom- passes the reasoning capability and internalized knowledge of LLMs, as well as tools, such as
\end{minipage}
\begin{wrapfigure}{r}{0.6\textwidth}
    \centering
    \vspace{10pt}
    \includegraphics[width=0.55\textwidth]{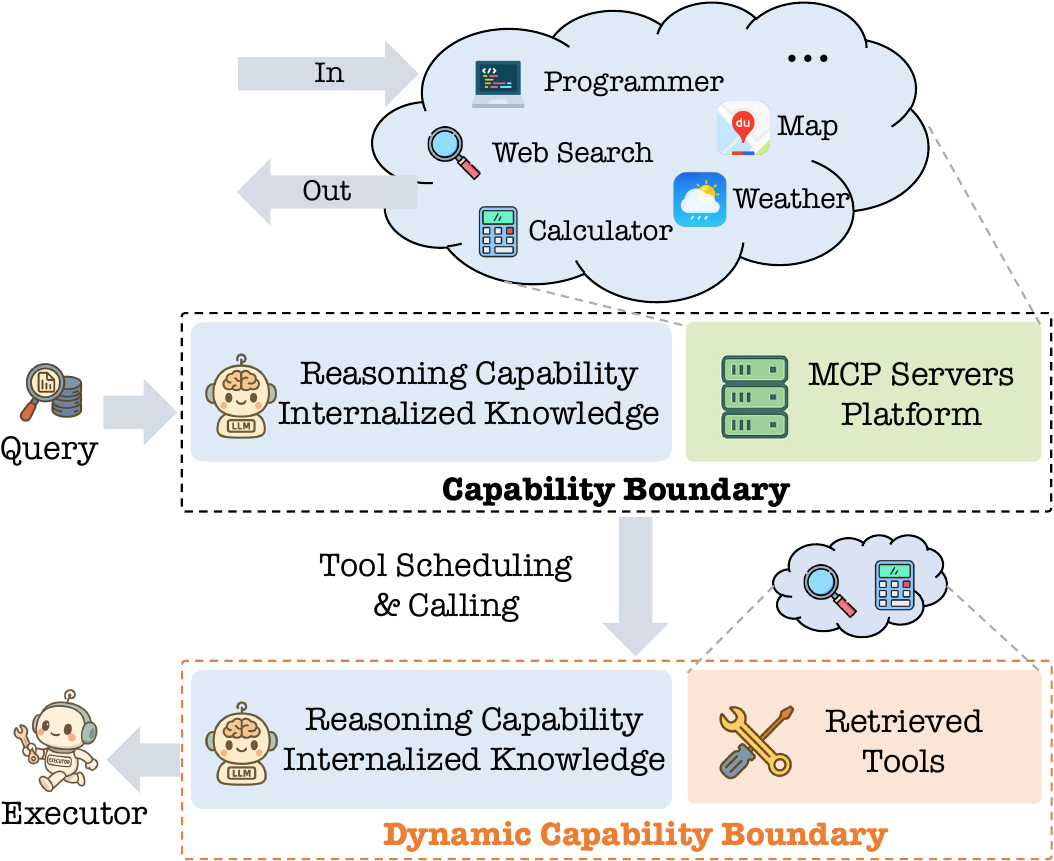}
    \caption{An illustration of the dynamic capability boundary of our search paradigm.}
    \label{plan_pic_dynamic}
\end{wrapfigure}
web search, calculators, and programmers. Typically, in the absence of these tools, an LLM
merely follows the traditional ``text input–and–text output'' paradigm inherent to conventional language model-based search paradigms; however, the integration of these tools, which enables functionalities such as web search, computation, and programming, substantially enhances the AI search process.
With the boundary defined, the AI search system can generate plans tailored to the input query. Specifically, the Planner constructs a DAG in which each node represents an individual sub-task (i.e., a single invocation of a tool), and the edges capture the dependency relationships between two nodes. This framework ensures that any task is executed only after all its prerequisite tasks have been completed.

However, a notable challenge arises as the portfolio of available tool APIs grows geometrically over time, eventually exceeding the representational capacity of a static capability boundary. To address the above challenge, AI Search paradigm introduces a novel concept termed the \emph{Dynamic Capability Boundary} within the task planner phase. As shown in Fig.~\ref{plan_pic_dynamic}, the AI search system leverages an LLM to process the input query and, within a short time, select a potential subset of tools. Given the selected tool subset, the AI search system combines it with the reasoning ability and internalized knowledge of the LLM to constitute the new dynamic capability boundary. Based on our extensive practical experience, the dynamic boundary typically consists of a relatively small set of tools, around a dozen tools, which proves sufficient to carry out task planning.


\subsubsection{Refinement of Tool API Documents}\label{Refined_Tool_API_Documents}
Considering the critical importance of precise tool description within AI Search paradigm, employing tool learning techniques to refine these documents represents an effective approach~\citep{mialon2023augmented,qin2023tool}. 
Specifically, tool learning has emerged as a powerful approach for enhancing the ability of LLMs to tackle complex, real-world tasks~\citep{schick2024toolformer,qu2025tool}. By interacting with tools such as calculators, search engines, APIs, and web environments, LLMs could overcome key limitations arising from their fixed pretraining data and inherently passive text-in/text-out interface~\citep{nakano2021webgpt,qin2023webcpm,m2024augmenting}. This integration allows LLMs to acquire real-time information, perform dynamic actions, and extend their utility beyond traditional language modeling~\citep{zhuang2023toolchain,wang2024llms}. 
To facilitate such interactions, LLMs are typically equipped with tool documentation that serves as contextual input, describing each tool’s functionalities, usage methods, and applicable scenarios~\citep{shen2024hugginggpt,song2023restgpt,xu2023tool}.
The effectiveness of LLMs in leveraging external tools largely depends on the clarity, structure, and accuracy of the accompanying API documentation. While traditional tool documents are generally written for human developers, they often fall short of meeting the specific needs of LLMs by containing ambiguities, redundancies, or missing information, thereby impeding the accurate interpretation of tool functionalities and task execution~\citep{chemero2023llms,yuan2024easytool}. Given these limitations, a key question arises: How can we construct tool documentation that is better aligned with the needs of the Planner?

\begin{wrapfigure}{r}{0.55\textwidth}
    \centering
    \vspace{-5pt}
    \includegraphics[width=0.5\textwidth]{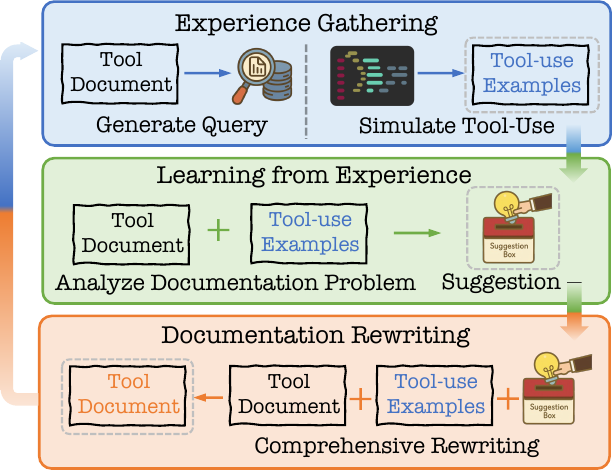}
    \caption{The pipeline of our proposed self-driven iterative refinement framework, which leverages the interactions between LLMs and external tools, as well as the feedback generated during these interactions, to progressively optimize tool documents}
    \label{exe_pic_draft}
    \vspace{-5pt}
\end{wrapfigure}

To enhance the quality of tool documentation, the AI search utilizes an iterative refinement method \emph{DRAFT}, which leverages the interactions between LLMs and external tools, as well as the feedback generated during these interactions, to progressively optimize tool documents~\citep{quexploration}. As illustrated in Fig.~\ref{exe_pic_draft}, DRAFT involves three iterative phases: \emph{Experience Gathering}, \emph{Learning from Experience}, and \emph{Documentation Rewriting}. 
Specifically, DRAFT first systematically simulates diverse use-cases, encompassing typical interactions, edge cases, error scenarios, and parameter limits. Such exhaustive exploratory interactions uncover gaps and inaccuracies in existing tool descriptions.
Subsequently, DRAFT analyzes collected interaction data to identify discrepancies and ambiguities. Insights derived from practical tool interactions ensure that refinements align closely with actual usage contexts. This analytical process yields targeted recommendations aimed at correcting inaccuracies, clarifying ambiguities, and eliminating redundancy.
Eventually, DRAFT integrates these recommendations, synthesizing refined tool descriptions specifically optimized for effective interpretation by LLMs. The method incorporates strategies that promote diverse exploratory coverage and adaptive termination criteria to prevent excessive iterations, ensuring concise yet comprehensive documentation. 
Next, we will describe the algorithm details of our proposed method as follows.


\textbf{Experience Gathering.} In the first phase, DRAFT leverages an explorer to simulate plausible scenarios for tool use, similar to how humans experiment with unfamiliar tools when manuals are unclear. 
Specifically, at the $i$-th iteration, $\mathcal{M}_{E}$ generates an exploration instance $e_i$ based on the current documentation $t_{i-1}$, the suggested direction $d_{i-1}$ from the rewriter $\mathcal{M}_{R}$, and prior history $\mathcal{H}_i = \{(e_j, r_j) \mid j<i\}$. The learning process could be formulated as 
\begin{equation}
    e_i = \mathcal{M}_E(t_{i-1}, d_{i-1}, \mathcal{H}_i), \label{Eq: explore agent}
\end{equation}
where $e_i$ comprises a query $e^q_i$ and its parameters $e^p_i$. The tool is then invoked to obtain the result returned by the tool, denoted as $r_i$. 

Given the complexity of tool usage, such as diverse parameter ranges and potential errors, it is essential to ensure exploration diversity. To this end, DRAFT implements a \emph{diversity-promoting strategy} to optimize this process. Specifically, the explorer will calculate the cosine similarity between the newly generated query $e^q_i$ and previous queries $e^q_j$~($j<i$).
A new instance is accepted only if:
\begin{equation}
\max_{j<i}\mathrm{sim}(\mathbf{e}^{q}_{i},\mathbf{e}^{q}_{j}) < \phi, \label{Eq:similarity}
\end{equation}
where $\phi$ is a predefined similarity threshold. If the constraint is violated, the explorer discards the instance and reflects on the redundancy~\citep{shinn2024reflexion}, iteratively refining the generated query until a sufficiently novel query is generated.

The diversity-promoting exploration strategy enables the explorer to engage with not only common tool functionalities but also less typical and edge-case scenarios. 
Such comprehensive interaction coverage helps reveal hidden flaws in the documentation, such as ambiguous instructions, incomplete descriptions, or misleading usage guidance. 
The exploration outcomes provide critical experiential data that form the foundation for later stages of analysis and document revision.

\textbf{Learning from Experience.} 
The second phase builds on the experiential data to refine the tool documentation. Here, DRAFT introduces an analyzer $\mathcal{M}_{A}$, which analyzes the exploration instance $e_i$, tool response $r_i$, current documentation $t_{i-1}$, and revision history $\mathcal{T}_i = \{ t_j \mid j < i \}$ to generate targeted revision suggestions $s_i$ as
\begin{equation}
    s_i = \mathcal{M}_A(t_{i-1}, e_{i}, r_{i}, \mathcal{T}_i). \label{Eq:suggest agent}
\end{equation}
To ensure relevance and quality, the analyzer is guided by criteria such as alignment with tool behavior, clarity, and informativeness without redundancy. By leveraging the revision history $\mathcal{T}_i$, it avoids repetitive suggestions and focuses on unresolved documentation gaps.

\textbf{Documentation Rewriting.} 
The final phase centers on document rewriting, aiming to improve the clarity, precision, and interpretability of tool documentation for LLMs. This phase not only incorporates feedback from prior exploration and revision but also proactively informs the next cycle of tool interaction.
To achieve this, DRAFT introduces a rewriting module, the rewriter $\mathcal{M}_R$, which consolidates multiple sources of information: (1) the exploration instances $e_i$ and corresponding tool outputs $r_i$ generated by the explorer $\mathcal{M}_E$; (2) the revision suggestions $s_i$ provided by the analyzer $\mathcal{M}_A$; and (3) the full rewrite history $\mathcal{T}_i$, which captures all previous versions of the documentation up to the current iteration. By considering the historical evolution of the documentation, the model can better preserve useful edits and avoid regressions.
Formally, the rewriting process could be defined as
\begin{equation}
    d_i, t_i = \mathcal{M}_R(t_{i-1}, e_i, r_i, s_i, \mathcal{T}_i), \label{Eq:rewrite agent}
\end{equation}
where $t_i$ represents the revised documentation at iteration $i$, and $d_i$ outlines recommended directions for further exploration. This closed-loop mechanism ensures that each iteration of rewriting is both context-aware and forward-looking, leading to increasingly robust and LLM-friendly documentation over time.

Recognizing that tool complexity varies, we adopt a \emph{tool-adaptive termination mechanism} to decide when to stop iterations. We measure the change $\Delta$ between consecutive versions using a combination of the word-match metric BLEU score~\citep{papineni-etal-2002-bleu} and the semantic-match metric cosine similarity, formulated as
\begin{equation}
\Delta = \frac{\mathrm{sim}(\mathbf{e}^{t}_{i},\mathbf{e}^{t}_{i-1})+\mathrm{BLEU}(t_i,t_{i-1})}{2}.
\label{Eq: early_break}
\end{equation}
If $\Delta$ exceeds a threshold $\tau$, we consider the documentation sufficiently refined and halt further revisions.
This tool-adaptive termination mechanism improves efficiency by avoiding redundant updates and ensures that the final documentation is both semantically and structurally aligned with LLM understanding.

DRAFT offers several key benefits that render it both practical and scalable for maintaining tool documentation. Fundamentally, DRAFT reimagines documentation refinement as a fully automated, feedback-driven process, thereby reducing the reliance on labor-intensive manual revisions. By leveraging LLMs’ iterative interactions with tools, it continuously identifies and rectifies discrepancies between the documented specifications and the LLMs’ actual behavior or interpretation. This iterative learning loop not only enhances clarity and usability but also guarantees that the documentation evolves concurrently with changes in the underlying tools. Moreover, as DRAFT operates through natural language prompts and responses, it provides strong interpretability and facilitates human review, auditing, and feedback, thereby ensuring an effective balance between automation and human-in-the-loop flexibility.

In this way, existing tool documentation can be systematically aligned with the specific requirements of the Planner, thereby laying a solid foundation for the subsequent planning process.

\subsubsection{Tool Clustering in MCP}
Effectively leveraging external tools in the AI search system requires a clear and fine-grained understanding of their functional characteristics. Existing categorizations of tool APIs, particularly in current MCP platforms, are often overly broad and fail to reflect task-specific functionalities. As a consequence, tools with distinct purposes may be grouped together, while those serving similar functions remain unlinked. This imprecise classification increases the complexity of task execution and reduces system reliability. In practical scenarios, when a tool fails during execution, the system often lacks access to alternative tools with equivalent capabilities. This absence of functional redundancy leads to higher error rates and diminished robustness, particularly under conditions of task complexity or environmental uncertainty.

To address these limitations, the AI search system concentrates on the automated categorization of tool APIs based on detailed functional similarities. Given the refined tool API document $D_{tool_i}$ introduced in Section~\ref{Refined_Tool_API_Documents}, the AI search system leverages LLMs to generate concise functional descriptions $desc_{tool_i}$ of tool APIs. 
These descriptions are subsequently transformed into high-dimensional semantic embeddings $e_{tool_i}$ using a text embedding model $\mathcal{M}_{e}$, which facilitates the identification of functionally coherent groups through rigorous clustering analyses. The above process can be formulated as
\begin{align}
    desc_{tool_i} = \mathcal{LLM}(D_{tool_i}),
\end{align}
\begin{align}
    e_{tool_i} = \mathcal{M}_{e}(desc_{tool_i}).
\end{align}

To organize the tool embeddings into functionally coherent groups, the AI search system uses the $k$-means++ algorithm~\citep{arthur2006k}, which partitions the embedding space into $k$ distinct clusters. The optimization objective of the clustering process can be formulated as
\begin{equation}
    \arg\min_{\mathcal{T}} \sum_{i=1}^{k} \sum_{e_{tool_i} \in \mathcal{T}_i} \left\| e_{tool_i} - \frac{1}{|\mathcal{T}_i|} \sum_{e \in \mathcal{T}_i} e \right\|_2^2,
\end{equation}
where $k$ denotes the predefined number of clusters, and each cluster $\mathcal{T}_i$ aims to minimize intra-cluster variance by aligning elements with their respective centroids.
In this way, the tool APIs can be grouped based on semantic similarity in functionality, regardless of their original grouping within the MCP platform\footnote{https://www.mcpworld.com/ai/mcp}. 

Such functional clustering is essential for ensuring system resilience. When a tool API fails during execution, having pre-grouped alternatives allows for immediate substitution with a functionally similar tool API. By implementing automated categorization, the system reduces reliance on manual annotations while enhancing the scalability and precision of tool API organization. 
Additionally, the AI search system refers to each cluster as a \emph{toolkit}, representing a functionally unified set of tool APIs capable of addressing a specific class of problems, such as \emph{search toolkit}, \emph{location toolkit}, \emph{weather toolkit}, \emph{programmer toolkit}, et al.
For instance, in the \emph{search toolkit}, there are \emph{Baidu AI Search}, \emph{ArXiv MCP Server}, \emph{Perplexity MCP Server}, \emph{OpenAI Web Search MCP}, et al.
It is imperative to emphasize that the \emph{search toolkit} contributes substantial practical value by ensuring resilient, diversified, and timely access to information. This enhanced functionality not only streamlines the search process but also significantly improves the operational efficiency and robustness of AI search systems.


\subsubsection{Query-Oriented Tool Retrieval}
While the AI search system has tailored the tool documentation to meet the specific needs of the Planner, interaction with a vast array of tools is often indispensable in real-world applications. In such settings, feeding all tool descriptions into an LLM is neither feasible nor efficient, due to input length restrictions and latency concerns. This scalability challenge highlights the necessity of a robust tool retrieval mechanism, one that can select the most suitable tools based on the task at hand, enabling LLMs to operate effectively in complex, tool-rich environments.
The AI search system initially employs an LLM-based dual-tower retrieval model for tool retrieval, which consists of an \emph{LLM-based Prompt Encoder} and an \emph{LLM-based Tool Encoder}. Specifically, the LLM-based prompt encoder extracts the user features and input prompts and generates the user embeddings. Meanwhile, the LLM-based tool encoder encodes the tool description, including the tool name, the tool usage, and the tool invoked queries, and produces the tool encodings. Subsequently, the AI search system computes the cosine similarity between the resulting user and tool embeddings and performs matching based on the similarity scores to retrieve the corresponding tools.

However, the above LLM-based tool retrieval strategies, typically focusing on semantic matching between queries and individual tools~\citep{xiong2020approximate,hofstatter2021efficiently,gao2021unsupervised,izacard2021unsupervised,ma2022pre}, inadequately address complex queries that necessitate collaborative tool use, resulting in incomplete or suboptimal outcomes.
For example, a query like calculating the value of 5 ounces of gold plus 1 million AMZN stocks in CNY requires tools for gold pricing, stock valuation, and currency conversion. If any are missing, the answer is incomplete. Traditional dense retrieval methods, focused on semantic similarity, often retrieve redundant tools (e.g., multiple stock tools) while omitting others, failing to capture the need for tool collaboration. This underscores the often-overlooked importance of completeness in tool retrieval.
Considering these factors, a central challenge arises: how can we design a retriever that selects a complete and task-relevant set of tools to support the Planner effectively?

To address the above challenge, the AI search system leverages a PLM-enhanced retrieval method \emph{COLT}, which integrates the semantic and collaborative dimensions of tool functionality~\citep{qu2024towards}. Initially, semantic relationships between queries and tools are established through representation learning techniques based on pre-trained language models. Subsequently, COLT employs graph-based learning methods to capture and refine collaborative interactions among tools, queries, and task scenarios. By adopting dual-view graph structures that explicitly model these interactions, COLT effectively captures the high-order collaborative information between tools. Furthermore, to promote balanced retrieval across all relevant tools, the learning objective incorporates a list-wise multi-label loss, preventing over-reliance on individual tools. Next, we detail the proposed tool retrieval method as follows.

\textbf{Sematic Learning.} 
COLT first focuses on semantic representation learning through dense retrieval, a widely adopted approach in modern retrieval systems~\citep{zhao2022dense, guo2022semantic}. 
It employs a bi-encoder architecture powered by PLMs, such as BERT~\citep{kenton2019bert}, to independently encode both the query $q$ and each tool description $t$ into dense vector embeddings. These embeddings, obtained via mean pooling over the final layer outputs, are then compared using cosine similarity to estimate their initial semantic relevance.
The training process is performed using the InfoNCE loss~\citep{gutmann2010noise,xiong2020approximate}, which contrasts positive tool-query pairs against negatives:
\begin{equation}
    -\log \frac{e^{\mathrm{sim}(q, t^+)}}{e^{\mathrm{sim}(q, t^+)} + \sum_{j=1}^{k} e^{\mathrm{sim}(q, t_{j}^{-})}}, \label{Eq.:infonce}
\end{equation}
where $t^+$ is the relevant tool, and $\{t_{1}^{-}, \cdots, t_{k}^{-}\}$ is a set of $k$ irrelevant tools.
In this way, COLT is trained to learn effective semantic representations. However, such retrieval alone often fails to handle complex, multifaceted queries, limiting the completeness of tool selection.

\textbf{Collaborative Learning.}
To move beyond simple semantic similarity and support completeness-oriented tool retrieval, it is crucial to model the collaborative relationships among tools.
COLT reinterprets the ground-truth tool set for a given query as a ``scene'', a conceptual unit that encapsulates the coordinated use of tools necessary to accomplish the underlying task.
Consider the query ``I want to travel to Paris." Rather than seeking a single response, the query implies a composite intent involving transportation, accommodation, weather forecasting, and tourist information. Addressing this multifaceted need requires retrieving a cohesive set of tools that operate collaboratively within the travel planning scene.
To model these intricate relationships, COLT constructs three bipartite graphs that jointly encode the associations among queries, tools, and scenes: \emph{query-scene}, \emph{query-tool} and \emph{scene-tool}.
Building on the semantic representations from the first stage and the three constructed bipartite graphs, we propose a dual-view graph collaborative learning framework that captures both direct query-tool relevance and higher-order tool interactions within scenes. This dual perspective improves retrieval completeness by modeling tool cooperation more effectively.
To align collaborative signals across views, COLT uses a cross-view contrastive loss. Additionally, a list-wise multi-label loss is introduced to ensure balanced retrieval from the full ground-truth set, avoiding overemphasis on individual tools.

With the cooperation of LLM-based tool retrieval and COLT, the AI search system could effectively retrieve a functionally complete set of tools from the entire tool repository in response to a given user query. This ensures that the downstream Planner is equipped with all the necessary tool capabilities, thereby facilitating more coherent and comprehensive planning and execution.

\subsection{DAG-based Task Planning}
Given the input query and the tool APIs retrieved accordingly, conventional approaches involve directly using these tools to obtain the required information and generate an answer. However, when the query is excessively complex and necessitates reasoning, this paradigm always fails to achieve the task through reasoning and multi-step tool invocation. In response, AI Search paradigm proposes a dynamic reasoning framework based on a DAG, effectively addressing the challenges associated with complex queries in the AI search system. 

Specifically, the AI search system leverages a candidate tool API set $\mathcal{T}$ and the user query $q$ as contextual input of the Planner, in a single inference pass, and generates a task graph  
$G=(V,E)$, where $V$ is the set of vertices (\ie, sub‑tasks) and $E$ refers to the set of dependency edges.  
Each vertex $v\in V$ represents an atomic and schedulable sub-task, which may either bind to an external tool or be fulfilled by the Executor using local LLM computation.
For uniform specification, every vertex is equipped with the interface signature described as $\bigl\langle \text{arg}(v), \ \text{ret}(v) \bigr\rangle$, where $\text{arg}(v)$ declares the parameter slots that must be instantiated before execution of the sub-task, and $\text{ret}(v)$ describes the structure of the data returned afterwards. If the sub‑task requires an external tool, the binding is recorded as $\tau(v)\in\mathcal{T}$; otherwise, $\tau(v)=\varnothing$ and the computation is performed locally.

Overall, the planning process can be formulated as the following mapping:
\begin{align}
    \Phi : (q,\mathcal{T}) \;\longrightarrow\; G ,
\end{align}
which subjects to two adequacy conditions:
\begin{itemize}
    \item \emph{Atomic Sub-Tasks.} For every vertex $v$ \emph{at most one} tool type $\tau(v)$ can be bound, although that tool may be invoked multiple times inside the sub‑task.  
    \item \emph{Sequential Dependency.} Second, the transitive closure of $G$ must capture all sequential dependencies inherent in the semantic relations required to address $q$, thereby ensuring completeness while minimizing inter‑task coupling.
\end{itemize}

To elicit a well‑structured DAG $G$ in a single forward pass, the AI search system employs a \emph{chain‑of‑thought $\rightarrow$ structured‑sketch} prompting strategy.  
Concretely, the Planner is first encouraged to decompose $q$ implicitly through reasoning, annotating latent variables and dependencies. Then, the resulting draft of the reasoning is reorganized into a JSON standard format for DAG via instruction prompting.
Unlike iterative ReAct-style reasoning, this approach produces a machine-readable, verifiable global plan simultaneously, substantially reducing token consumption and simplifying downstream parsing.

Once the task DAG $G$ is established, the AI search system schedules execution by topological depth $\ell(\cdot)$.  
For each layer $V_d=\{v\mid\ell(v)=d\}$, all vertices are carried out in parallel.
If $\tau(v_i)\neq\varnothing$ the corresponding tool is invoked one or more times, otherwise the sub-task is performed by the Executor individually. After executing, the resulting $\text{ret}(v_i)$ is then propagated along edge $(v_i,v_j)$ to satisfy $\text{arg}(v_j)$.  
In the event of a vertex failure, the Master forwards the failure signal to the Planner.
Subsequently, the Planner initiates a localized rollback and re-planning on the affected subgraph while preserving the remainder of the DAG $G$. 
This method enhances robustness without incurring the cost of a global restart.

In this way, DAG-based task planning externalizes multi-step reasoning into a structured task graph, achieving minimal context length, layer-wise parallelism, and full traceability. This design provides a principled, modular backbone for subsequent reflection mechanisms and reinforcement‑learning optimization, which will be discussed in the remaining sections.

\subsection{Master-Guided Re-Action}
Real-time monitoring of sub-task execution and the evaluation of intermediate results, coupled with the ability to reflect, re-plan, and re-act based on those outcomes, is critical for AI Search paradigm. Therefore, the AI search system leverages a \emph{Mater-guided reflection, re-planning and re-action} mechanism.

Given an input query, the Planner initially constructs a DAG, after which the Executor carries out each task node until the final result is produced. Throughout this process, the Master continuously evaluates both the execution status of the sub-tasks and the completeness of their outcomes, engaging in reflection and dynamically updating the DAG as necessary. 
Specifically, in the AI search system, the Planner initially constructs a DAG based on the provided query, thereby outlining a series of sub-tasks to be executed. Subsequently, the Executor carries out each prescribed sub-task according to the DAG. If any sub-task fails to execute, the Master directs the Planner to re-plan either the affected portion of the DAG or the entire DAG. Conversely, upon the completion of each stage, whether the subtasks are executed individually or concurrently, the Master assesses the execution results of each subtask. If the results are found to be incomplete, the Master guides the Planner to augment the existing DAG by incorporating additional downstream sub-task nodes to achieve complete results.
\subsection{Planner Optimization with RL Strategy}\label{ssec:how_optimize_planner}
In practice, high-quality, accurately annotated data for supervised fine-tuning (SFT) presents substantial challenges, primarily because of the significant cost associated with manual annotation. Furthermore, AI search paradigm comprises four specialized agents (i.e., Master, Planner, Executor, and Writer) that collaboratively process queries to generate the final output. Consequently, optimizing the Planner in isolation may prove insufficient and could lead to suboptimal performance of the overall system. To address this issue, a promising strategy is to employ reinforcement learning (RL) to optimize the Planner in the context of multi-agent cooperation. Specifically, the proposed AI search system introduces a rule-based reward function that consists of four components: (1) \emph{Final Answer Rewards}, (2) \emph{User Rewards}, (3) \emph{Formatting Rewards}, and (4) \emph{Intermediate Execution Rewards}, as detailed below.
\begin{itemize}
    \item \textbf{Final Answer Rewards.} The primary objective of the Planner is to decompose the task to obtain the correct final answer. Accordingly, the final answer reward could be defined as
    \begin{equation}
        \mathcal{R}_{Answer} = \alpha_{Answer} \cdot \mathds{1}[Answer =\mathrm{G}_{Answer} ],
    \end{equation}
    where $\alpha_{Answer} > 0$ is a scaling factor, and $Answer$ denotes the final answer produced by the multi-agent workflow. $\mathrm{G}_{Answer}$ is the ground-truth answer.

    \item \textbf{User Feedback Rewards.} User feedback serves as an effective signal, enabling the Planner to further adjust the decomposition of sub-tasks. Consequently, the reward of user feedback could be formulated as
    \begin{equation}
    \mathcal{R}_{Feedback} =  
        \begin{cases}
            \alpha_{Feedback}, & \text{if users accept the sub-tasks;} \\
            \beta_{Feedback}, & \text{if users reject the sub-tasks,} 
        \end{cases}
    \end{equation}
    where $\alpha_{Feedback} > 0$ is the reward when the sub-tasks are accepted by users, while $\beta_{Feedback} < 0$ is the penalty when they are rejected. $\alpha_{Feedback}$ and $\beta_{Feedback}$ are both scaling factors.

    \item \textbf{Formatting Rewards.} To ensure the generated output adheres to the desired format, the formatting reward could be defined as
    \begin{equation}
        \mathcal{R}_{Format} =  
        \begin{cases}
            \alpha_{Format}, & \text{if output is correctly formatted;} \\
            \beta_{Format}, & \text{if output is incorrectly formatted,} 
        \end{cases}
    \end{equation}
    where $\alpha_{Format} > 0$ represents the reward for correct formatting, whereas $\beta_{Format} < 0$ denotes the penalty for formatting errors. $\alpha_{Format}$ and $\beta_{Format}$ are two scaling factors.
    
    \item \textbf{Intermediate Execution Rewards.} To ensure the executability of the decomposed sub-tasks, the intermediate execution reward could be defined as
    \begin{equation}
        \mathcal{R}_{Execution} = \frac{\alpha_{Execution}}{m}\cdot\sum_{i=1}^{m} f_{Executor}(task_i),
    \end{equation}
    where $\alpha_{Execution} > 0$ is a scaling factor, and $m$ is the total number of sub-tasks. Moreover, $task_i$ denotes the $i$-th sub-task, and $f_{Executor}(task_i) \in [0,1]$ represents the execution success score of each sub-task.

    \item \textbf{Overall Reward and Objective Function.} The overall reward could be formulated as the sum of all the individual reward components mentioned above
    \begin{equation}
        \mathcal{R}_{All} = \mathcal{R}_{Answer} + \mathcal{R}_{Feedback} + \mathcal{R}_{Format} + \mathcal{R}_{Execution}.
    \label{eq:reward_all}
    \end{equation}
    \end{itemize}

Given the above rewards, the AI search system leverages Group Relative Policy Optimization (GRPO)~\citep{shao2024deepseekmathpushinglimitsmathematical} to optimize the Planner. Specifically, for a given complex query $q$, the AI search system samples a set of outputs $o = \{o_1, o_2, \cdots, o_N\}$ from the previous planner policy model $\pi_{\theta_\text{old}}$. Corresponding to these outputs, it computes a set of rewards $r = \{r_1, r_2, \cdots, r_N\}$ based on the rule-defined reward function (Eq.~\ref{eq:reward_all}). 
The advantage for each token is defined as the normalized reward as
\begin{equation}
    \hat{A}_{i,t}=\widetilde{r}_i=\frac{r_i-\mathrm{mean}(\mathbf{r})}{\mathrm{std}(\mathbf{r})}.
\end{equation}

The AI search system optimizes the policy by maximizing the following objective function as

\begin{equation}
    \begin{split}
    \mathcal{J}_{GRPO}(\theta) &= \mathbb{E}[q \sim P(Q), \{o_i\}_{i=1}^{G} \sim \pi_{\theta_{old}}(O|q)] \\
    & \frac{1}{G}\sum_{i=1}^{G}\frac{1}{|o_i|}\sum_{t=1}^{|o_i|} \left\{\gamma - \beta\mathbb{D}_{KL}\left[\pi_{\theta}||\pi_{ref}\right]\right\},
    \end{split}
\end{equation}
where $\gamma = \min\left[\lambda  \hat{A}_{i,t}, \text{clip}\left(\lambda , 1-\varepsilon, 1+\varepsilon\right)\hat{A}_{i,t}\right] $ , and $\lambda  = \frac{\pi_{\theta}(o_{i,t}|q, o_{i,<t})}{\pi_{\theta_{old}}(o_{i,t}|q, o_{i,<t})}$.  Here, $\pi_{\theta}$ denotes the current planner policy model, and $o_{i,t}$ is the $t$-th token of the output sequence $o_i$. Additionally, $o_{i,<t}$ represents the tokens preceding $o_{i,t}$, and $\beta$ and $\varepsilon$ are hyper-parameters.

\section{Task Executor}
Following the meticulously designed task plan generated by the Planner, the AI search paradigm incorporates a task executor module, known as the Executor, which is responsible for invoking the appropriate tools in accordance with the prescribed plan and executing specific subtasks to produce the requisite outcomes for final answer generation. Among these tools, the web search functionality is a critical component and represents the most frequently utilized tool within the system. This need arises not only from queries that require information beyond the model’s parametric knowledge, such as retrieving current gold prices, but also from the imperative to provide query responses that are both timely and authoritative, thereby meeting key user satisfaction dimensions. The following section presents our methodological advancements aimed at enhancing the capability of the web search component.

\subsection{Background}


In traditional pre-trained language model (PLM)-based retrieval systems, task execution primarily involves retrieving and ranking a list of documents in response to a single search query, with the goal of meeting the immediate needs of the user. This approach typically transforms a user’s query into a ranked document list, from top to bottom, aiming to deliver the most relevant results efficiently. With the rapid development of LLMs, there has been growing interest among researchers to leverage them for document ranking. Currently, there are three main approaches for document ranking with LLMs based on their input, such as \emph{Pointwise}, \emph{Pairwise}, and \emph{Listwise}, which are introduced as follows.

\textbf{Pointwise}. These approaches utilize LLMs to independently assess the relevance of each candidate document with the given query. Formally, given a query $q$ and a retrieved document $\mathcal{D}_{i}$, we aim to instruct the LLM to generate a quantitative, comparable score for each query and candidate document pair, denoted as the basic labeling unit as $u(q, \mathcal{D}_{i})$. This score can be a graded label commonly used in human annotation or a logit value output by the model. Then, we apply the labeling unit to all the candidate documents and obtain the final rankings based on the sorted scores. For instance, some works directly prompt LLMs to generate whether the provided candidate document is relevant to the query and the documents are labeled based on the normalized likelihood of generating a ``\emph{yes}'' response \citep{nogueira2020document, Liang2022HolisticEO,sun2023instruction}. The other method involves employing LLMs to generate a relevant query for each candidate document and then re-rank these documents based on the likelihood of generating the actual query \citep{Sachan2022ImprovingPR}. Another approach involves appending an output layer to the LLM's representation and then employing SFT with human-annotated samples for re-ranking tasks. This process produces a scalar score that reflects the relative query-document relevance~\citep{10.1145/3626772.3657951}.

\textbf{Pairwise}. These approaches use LLMs to compare the relevance of a pair of candidate documents with the given query. Formally, given a query $q$ and a pair of retrieved documents $(\mathcal{D}_{i}, \mathcal{D}_{j})$, we aim to ask the LLM to determine which document is more relevant to the given query $q$, denoted as the basic labeling unit $u(q, \mathcal{D}_{i}, \mathcal{D}_{j})$. There are two principal approaches leveraging the pairwise score unit: the all-pairs method and the sorting-based method \citep{Pradeep2021TheED, Qin2023LargeLM}. The all-pairs method generates all possible permutations of document pairs from the candidate documents set. Then, each pair is evaluated independently by the LLM, and a score function is then used to assign a score to each document based on the pairwise comparisons. The final ranking is determined by the total score each document receives. In contrast, the sorting-based method utilizes sorting algorithms, such as heap sort and bubble sort, to leverage efficient data structures for selectively comparing document pairs. This enables the most relevant documents to be quickly identified and placed at the top of the final ranking without the need for exhaustive comparison of all possible pairs. More advanced research, such as PRP-Graph \citep{Luo2024PRPGraphPR}, draws inspiration from real-world tournaments and incorporates graph structures into pairwise ranking to further improve efficiency.

\begin{wrapfigure}{r}{0.55\textwidth}
    \centering
    \vspace{-5pt}
    \includegraphics[width=0.52\textwidth]{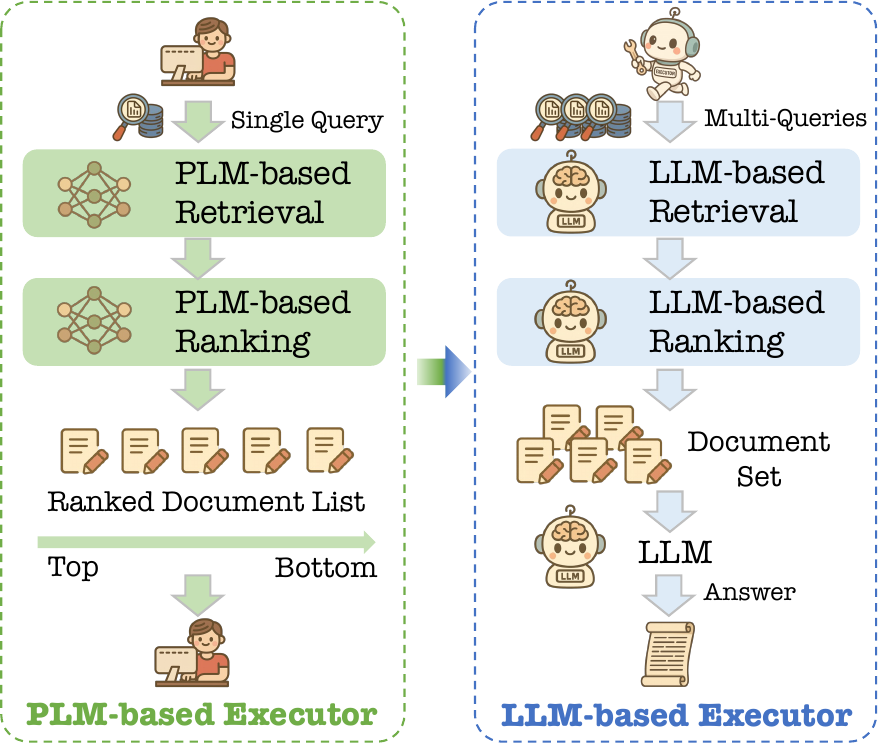}
    \caption{An illustration of the paradigm shift for the task Executor. The left panel shows the traditional task executor based on PLMs for web search, which primarily addresses a single query by transforming the web search task into a ranking problem and aligning the results with user preferences. The right panel illustrates the paradigm shift in the task execution for AI search. Our AI search system processes multiple queries by re-framing the traditional ranking problem as a recall problem, shifting the objective from aligning user preferences to aligning LLM preferences.}
    \label{PLM_to_LLM}
    \vspace{-10pt}
\end{wrapfigure}
\textbf{Listwise}. These approaches utilize LLMs to directly rank multiple documents within a single window based on their relevance to the given query~\citep{Sun2023IsCG,liu2024sliding,Chen2024TourRankUL}. Formally, given a query $q$ and a list of retrieved
documents $\{\mathcal{D}_{1}, \dots, \mathcal{D}_{m}\}$, we aim to ask the LLM to directly generate an ordered doc-list based
on the relevance to the given query $q$, denoted as the basic labeling unit $u(q, \{\mathcal{D}_{1}, \dots, \mathcal{D}_{m}\})$. Since the Listwise method achieves a good trade-off between effectiveness and efficiency for ranking multiple documents at one time, they are considered as the preferred prompting strategy for LLM ranking. 
In the following sections, we will detail two Listwise-based ranking methods in practice: RankGPT \citep{Sun2023IsCG} and TourRank \citep{Chen2024TourRankUL}.

While effective in straightforward search scenarios, these methods are still based on the traditional paradigm, lacking the flexibility and adaptability necessary to handle more complex interactions, such as multi-turn conversations, intent-driven query decomposition, and answer synthesis, which are increasingly important in modern AI applications.

To address these limitations, recent technological developments have paved the way for a fundamental shift in retrieval paradigms. With the rapid advancements in LLMs, there has been a significant paradigm shift in task execution strategies. As illustrated in Fig.~\ref{PLM_to_LLM}, our AI search system moves away from the traditional PLM-based executor towards an integrated approach that leverages both retrieval and reasoning capabilities of LLMs. Instead of solely ranking documents based on a single query, the system now processes multiple, complex queries, potentially across multiple rounds, using LLM-based retrieval and ranking modules. The goal shifts from merely producing a ranked list for a specific user need to recalling a comprehensive document set that can serve as a reference for the LLM. This set provides the contextual foundation for the LLM to generate high-quality, nuanced answers tailored to the user’s evolving intent.

Building upon this motivation, the AI search system focuses on designing a task executor that effectively harnesses LLM capabilities. In this section, we detail the design and motivation behind our task executor, focusing on three core perspectives: \emph{LLM Preference Alignment}, which ensures the system’s targets align with LLMs; \emph{Lightweight System}, aimed at processing multiple complex queries; and \emph{LLM-Augmented Features}, which enhances the retrieval process with LLM-augmented features.

\subsection{LLM Preference Alignment}
Traditional retrieval and ranking systems primarily rely on manually defined preferences and heuristics, which often fail to fully leverage the deep understanding and nuanced preferences of LLMs. As LLMs demonstrate significant advantages in comprehending complex semantics, handling multi-modal data, and performing sophisticated reasoning, the goal shifts from merely aligning with heuristic-guided preferences to aligning with the LLM-human co-evolutionary preferences. This transition enables the retrieval system to better facilitate high-quality answer generation by guiding the selection towards documents favored by the LLM. Ultimately, aligning with LLM preferences enables more intelligent, autonomous, and effective retrieval strategies, thereby improving the overall performance of the question-answering system.

To achieve this, the AI search system constructs data from three perspectives: \emph{LLM Labeling}, \emph{Reference Selection}, \emph{Generation Reward}. These perspectives guide the training process by capturing different aspects of LLM preferences. Moreover, the AI search system further aligns the preference into the LLM retrieval model through \emph{Distillation of LLM Ranking}, which helps the model internalize the learned preferences and improve its retrieval performance.

\subsubsection{LLM Labeling} 
Compared to manual labeling processes, LLMs can efficiently annotate vast datasets at scale, thereby substantially reducing both the temporal and human resource investments required for training retrieval and ranking models. Furthermore, LLMs deliver highly consistent annotations, mitigating the subjective biases and inter-annotator variability that commonly afflict human-generated labels. By adjusting prompt designs or fine-tuning for specific domains or tasks, LLMs exhibit exceptional flexibility in meeting diverse annotation requirements. As previously noted, the AI search system utilizes listwise methods for document annotation, as these approaches effectively strike a balance between effectiveness and efficiency. In this work, the AI search system primarily utilizes two methods as follows.: RankGPT \citep{Sun2023IsCG} and TourRank \citep{Chen2024TourRankUL}.

\textbf{RankGPT.} When LLMs come to ranking tasks, especially in the context of passage re-ranking, the limited context window of LLMs poses a significant challenge. Most LLMs have a maximum context length that restricts the number of passages they can process simultaneously. This limitation is particularly problematic in scenarios where a large number of candidate documents need to be ranked, such as in web search or large-scale IR systems.
To address this challenge, the AI search system utilizes a strategy called the sliding window approach RankGPT~\citep{Sun2023IsCG}. This method allows the efficient re-ranking of large candidate sets without exceeding the context window limitations of the LLM. By breaking down the ranking task into smaller, manageable segments, RankGPT ensures that the LLM can effectively process and rank each subset of documents while maintaining the overall ranking quality.

The sliding window strategy employed by RankGPT involves ranking a subset of candidate documents within a sliding window, starting from the bottom of the initial ranking list and moving forward. This approach ensures that each document is compared with multiple others across overlapping windows, maintaining the effectiveness of LLM ranking while remaining computationally feasible for long candidate lists. The process can be broken down into the following stages:

\begin{itemize}
    \item \textbf{Initialization.} Given the retrieved set of \( M \) candidate documents from the retrieval model,
    the sliding window strategy begins with the last \( w \) documents, i.e., from the \( (M-w+1) \)-th to the \( M \)-th passage. These documents are input to the LLM, which returns a ranked order for this subset. The choice of \( w \) (window size) is crucial as it determines the number of documents processed in each iteration. A larger window size allows the LLM to consider more documents at once, potentially improving the ranking quality, but also increases the computational load and the risk of exceeding the context window limitations.
    \item \textbf{Sliding.} Once the initial window is ranked, the window slides backward by \( s \) documents. The new window now covers documents from \( (M-w-s+1) \) to \( (M-s) \). The LLM is again used to rank this subset, which may overlap with the previous window. The step size \( s \) controls the overlap between consecutive windows. A smaller step size increases the overlap, allowing the LLM to compare more documents across windows, but also increases the number of iterations required to cover all documents.
    \item \textbf{Iteration.} The sliding and re-ranking process is repeated until the window reaches the beginning of the passage list and all documents have been covered. This iterative approach ensures that each document is ranked multiple times, allowing the LLM to provide more nuanced and consistent ranking signals across the entire candidate set.
\end{itemize}

\begin{figure*}[!t]
    \centering
    \includegraphics[width=0.95\textwidth]{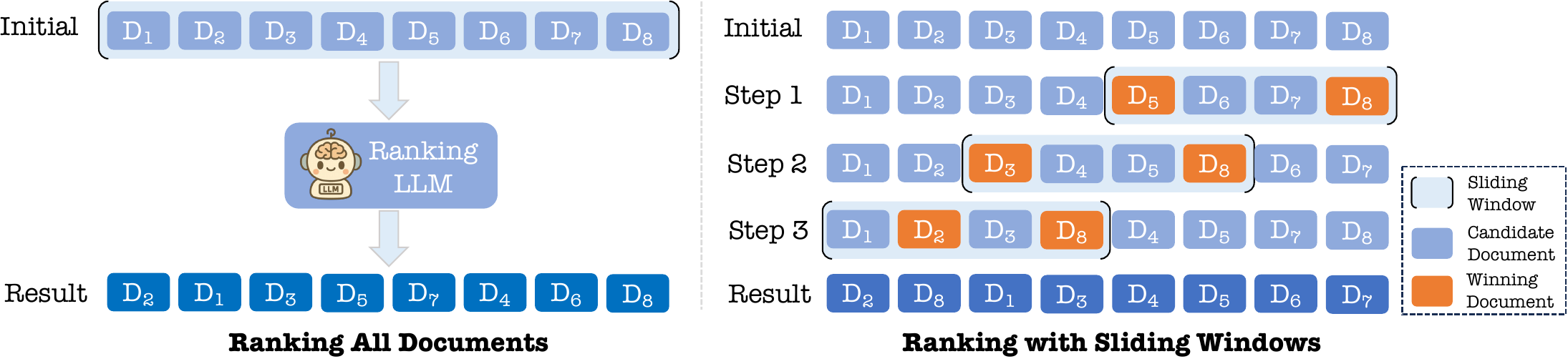}
    \caption{\centering The illustration of ranking all documents and ranking with sliding windows.}
    \label{fig:overall_framework}
\end{figure*}



In conclusion, the sliding window strategy employed by RankGPT effectively addresses the context window limitations of LLMs, making it possible to leverage their powerful ranking capabilities for large-scale information retrieval tasks. This approach not only improves the efficiency and effectiveness of passage re-ranking but also offers a scalable solution for various IR applications.

\textbf{TourRank.} LLMs have shown great potential in zero-shot document ranking. However, existing LLM-based ranking methods face several limitations. First, the maximum context length of LLMs restricts the number of documents that can be processed simultaneously, making it challenging to rank a large number of documents at once. Second, the ranking results are highly sensitive to the order in which documents are presented to the LLM, leading to inconsistent outcomes. Third, achieving a balance between computational cost and ranking performance remains a significant challenge. To address these issues, the AI search system introduces a novel approach inspired by the structure of sports tournaments, named TourRank~\citep{Chen2024TourRankUL}.

In sports tournaments, there are often multiple participants, which requires that their rankings be determined as fairly as possible within a short period of time. In many sports tournaments, such as the FIFA World Cup, a group stage format is often adopted. Top-seeded players or teams are distributed across different groups, and multiple group matches are conducted in parallel. This allows for the rapid selection of the top-ranked participants, such as the gold, silver, and bronze medalists.

Similarly, if we liken the document ranking process to a sports tournament, each document can be considered as a participant. By simulating the structure of sports tournaments, we can quickly obtain the ranking of the documents. Specifically, suppose we have $M$ documents to rank. A basic tournament proceeds as follows:

\begin{itemize}
    \item \textbf{Group and Select.} Given a set of $M$ candidate documents, TourRank begins by dividing them into several smaller groups. Each group contains a manageable number of documents that fit within the context window of the LLM. For example, if there exist 100 documents to rank, they might be divided into groups of 20 documents each. Within each group, the LLM evaluates the relevance of the documents to the query and selects the top $N$ most relevant documents to advance to the next stage. This process is analogous to the group stage in sports tournaments, where teams compete within their respective groups to qualify for the next round.
    
    \item \textbf{Progressive Rounds.} After the initial grouping and selection, the winners from each group are pooled together and re-divided into new groups for the next round. This hierarchical, multi-stage structure continues iteratively, with each round reducing the number of documents and selecting the most relevant ones. For instance, when starting with 100 documents divided into 10 groups of 10, the first round selects 5 documents from each group, thereby reducing the total to 50 documents. These 50 documents are then divided into 5 groups of 10, and the process repeats. Each document earns 1 point whenever it is selected to advance to the next stage.
    
    \item \textbf{Points System and Multiple Rounds.} The points system is a crucial component of TourRank. Since the scores obtained from a single tournament can be coarse and may contain randomness, TourRank conducts $R$ rounds of tournaments in parallel. Each document's total score is computed by summing its scores from all rounds, denoted as
    \begin{equation}\label{eq:tourrank}
        \text{score}_{d_i} = \sum_{r=1}^{R} \text{score}_{d_i}^{r},
    \end{equation}
    where $R$ is the total number of rounds, and $\text{score}_{d_i}^{r}$ is the score assigned to document $d_i$ in round $r$. This approach ensures that the final ranking is more fine-grained and robust, as it aggregates the results from multiple independent tournaments.
\end{itemize}

\begin{figure*}[!t]
    \centering
    \includegraphics[width=0.91\textwidth]{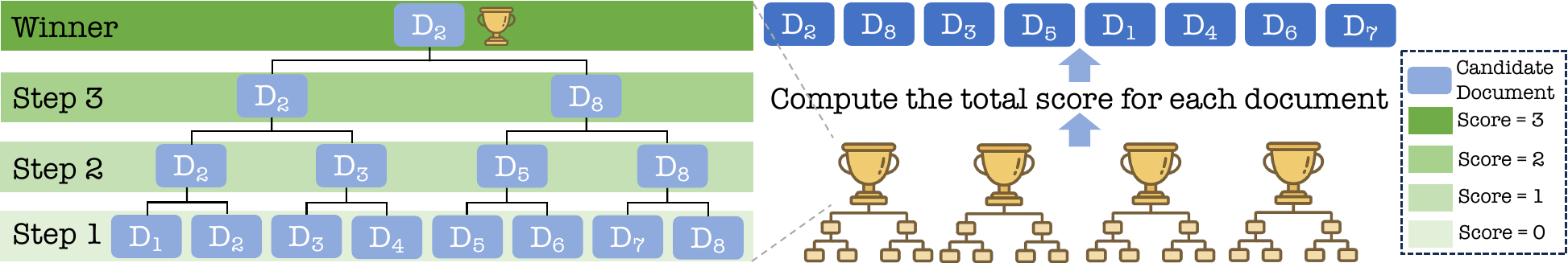}
    \caption{\centering The illustration of ranking with a tournament strategy.}
    \label{fig:overall_framework}
\end{figure*}

The tournament structure of TourRank significantly enhances its efficiency and scalability. By allowing parallel processing within and across groups, TourRank reduces the overall ranking latency. Specifically, the time complexity of TourRank is $\mathcal{O}(K-1)$, where $K$ represents the number of stages in a single tournament. This is a substantial improvement over serial sliding window methods, which have a time complexity of $\mathcal{O}\left(\frac{N-\omega}{s}\right)$. The parallel nature of TourRank makes it particularly suitable for large-scale ranking tasks, where the number of candidate documents can be very large.

\subsubsection{Reference Selection}
This section highlights the importance of selecting high-quality references in indirectly aligning the LLM preference, ensuring that retrieved documents can enhance the generation of answers. For a complex query, the system retrieves multiple candidate references, which are then incorporated as contextual input to the LLM during the synthesis of an answer. The model’s generated response is explicitly linked to supporting references (e.g., ``11 of 50 state names are derived from source [1]; Texas is named after a Native American tribe as indicated in source [3], ...''), thereby ensuring that the answer is grounded in verifiable sources. This explicit citation mechanism not only enhances the transparency and trustworthiness of the answer but also facilitates subsequent verification and validation by users and downstream systems. Furthermore, a feedback loop enables iterative refinement of retrieval strategies based on observed LLM usage and outcomes, which in turn optimizes future reference selection for analogous queries. When an answer is generated, the AI search system collects signals related to user satisfaction, answer accuracy, and the relevance of the cited references. These signals are then used to adjust retrieval parameters, ranking functions, or even the candidate source pool in subsequent retrieval cycles. For instance, if the LLM consistently demonstrates a preference for certain types of references (e.g., recent publications, authoritative websites, or data-rich documents), the retrieval system can be adjusted to prioritize these sources for similar future queries.

High-quality reference selection, guided and refined by LLM feedback, not only grounds generated answers in reliable sources but also establishes a virtuous cycle of continuous improvement. This methodology is crucial for building trustworthy, adaptive, and high-performing LLM-based question-answering and information retrieval systems.

\begin{figure*}[!t]
    \centering
    \includegraphics[width=0.91\textwidth]{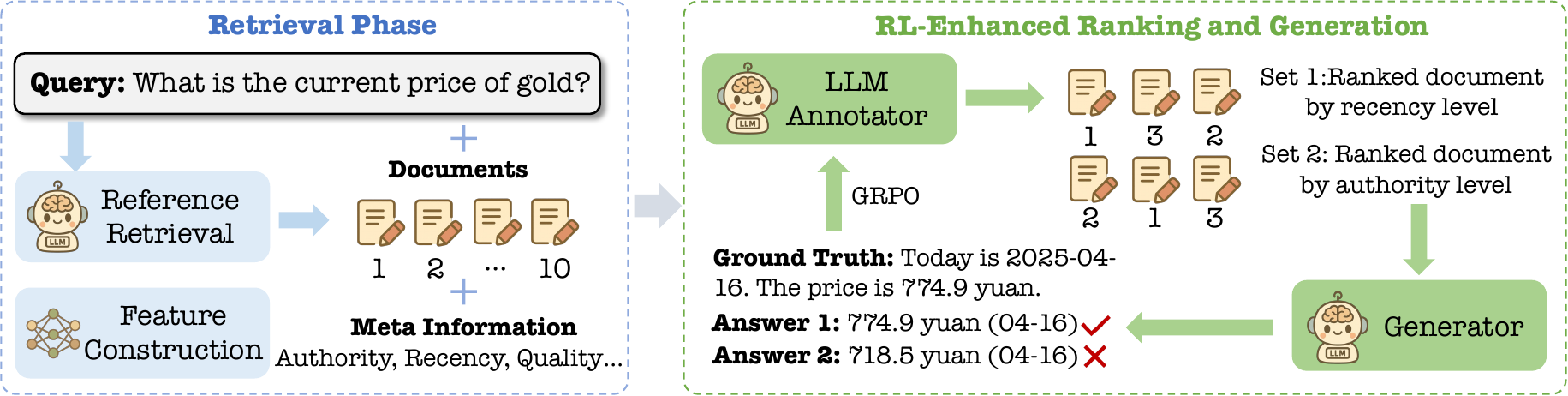}
    \caption{\centering The illustration of generation reward.}
    \label{fig:overall_framework}
\end{figure*}

\subsubsection{Generation Reward}
Reference selection indirectly aligns the ranker with LLM preferences, whereas generation rewards achieve this alignment directly. Traditional reference retrieval systems have predominantly relied on supervised learning to optimize the relevance of results. However, the integration of LLMs introduces new opportunities for dynamic and intelligent feedback loops. A promising approach leverages feedback from LLM-generated answers to inform and refine retrieval ranking via reinforcement learning. In this framework, LLM responses serve as direct feedback that guides the optimization of both retrieval and ranking processes.

For each query and its candidate documents, accompanied by metadata such as site name, publication time, and quality score, the ranker applies multiple policies that emphasize different feature combinations (e.g., recency, quality, relevance, and authority). The resulting document rankings are then evaluated using a frozen generator that produces multiple candidate answers. These generated answers are compared with ground truth responses, and a reinforcement reward is assigned to the ranker based on performance. This reward-driven methodology facilitates dynamic policy selection and enhanced document ranking, directly calibrated to generation outcomes. In implementing the reinforcement learning algorithm, the AI search system leverages GRPO. Incorporating generation rewards via reinforcement learning represents a significant advancement toward truly intelligent and adaptive reference retrieval systems. Rather than treating retrieval, ranking, and answer generation as isolated tasks, this integrated feedback loop directly optimizes for both user satisfaction and answer quality.

For example, given a query ``\emph{What is the current price of gold?}'', the AI search system retrieves relevant documents along with their metadata. The ranker may then select a policy, such as $\mathcal{P}_{1}: \{recency \succ quality \succ relevance \succ authority\}$, or $\mathcal{P}_{2}: \{quality \succ relevance \succ recency \succ authority\}$, to sort these documents accordingly. The sorted results are forwarded to the generator, and feedback from the generator is subsequently used to reward the ranker. This process facilitates document ranking that effectively integrates metadata, thereby optimizing overall retrieval performance.

\subsubsection{Distillation of LLM Ranking}
Consider a query $q$ and its corresponding final ranking of candidate documents $\{\mathcal{D}_{1}, \dots, \mathcal{D}_{m}\}$ obtained from a reasoning-LLM via the methods described above (i.e., \emph{LLM Labeling}, \emph{Reference Selection}, and \emph{Generation Reward}). The teacher-generated data are used to construct $m*(m-1)/2$ document pairs $(\mathcal{D}_{i}, \mathcal{D}_{j})$. Subsequently, an online ranker, which operates as a pointwise ranking LLM, assumes the role of the student model. This student is optimized using the RankNet loss, a pairwise loss function that measures the accuracy of the relative ordering between items, as defined by
\begin{equation}\label{eq:ranknet}
    \mathcal{L} = \sum_{i=1}^n \sum_{j=1}^n \mathbf{1}_{r_i^t < r_j^t} \log(1 + \exp(s_i^\mathrm{s} - s_j^\mathrm{s})),
\end{equation}
where $r_i^t$ and $r_j^t$ denote the teacher's ranking, and $s_i^\mathrm{s}$ and $s_j^\mathrm{s}$ represent the corresponding student scores. This distillation process transfers the deep ranking reasoning of the teacher LLM to a more efficient student model suitable for real-time deployment.


\subsection{Lightweight System}
Modern user queries are increasingly complex, multi-turn, and multifaceted, which poses significant challenges to traditional retrieval systems that are primarily optimized for simple keyword searches. Decomposing such queries into multiple sub-queries, managing iterative interactions, and integrating diverse search intents not only increases computational overhead but also augments system complexity. To overcome these challenges, it is essential to develop a lightweight and efficient retrieval system that leverages the deep comprehension and reasoning capabilities of LLMs. Such a system can streamline system architecture, reduce latency, and effectively manage large-scale, multifaceted queries, thereby enabling faster and more accurate information retrieval in dynamic, real-world scenarios. To achieve this objective, we propose a lightweight system designed to efficiently process massive natural language sub-queries. Specifically, AI Search paradigm represents a paradigm shift from the traditional hybrid retrieval methods—combining inverted indices with dense retrieval and subsequent PLM-based re-ranking—to a novel framework in which LLMs are directly employed to simplify retrieval and rank relevant documents. This transition not only simplifies the underlying architecture but also fully harnesses the deep contextual reasoning capabilities of LLMs.

\subsubsection{Lightweight Retrieval}
\begin{wrapfigure}{r}{0.5\textwidth}
    \centering
    \includegraphics[width=0.48\textwidth]{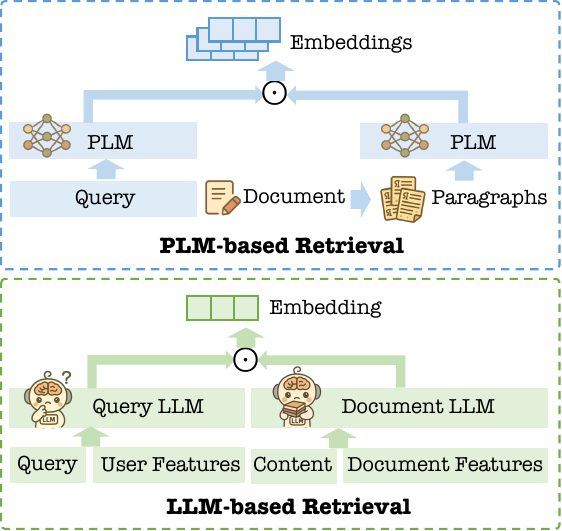}
    \caption{An illustration of the comparison of PLM-based Retrieval and LLM-based Retrieval.}
    \vspace{-5pt}
    \label{exe_retrieval_compare}
\end{wrapfigure}

In the evolution of retrieval systems, various approaches have emerged to enhance accuracy and personalization. Traditional methods initially employed PLMs to generate embeddings for queries and documents, thereby facilitating relevance-based retrieval. However, the increasing demand for a more nuanced understanding and multi-modal integration has spurred the development of advanced retrieval architectures based on LLMs. Fig.~\ref{exe_retrieval_compare} provides a visual representation of this progression, outlining the transition from the conventional PLM-based retrieval system to the LLM-driven retrieval system.

\textbf{PLM-based Retrieval.} Early literature on retrieval systems typically employed language models to generate embeddings for both queries and document paragraphs. In these systems, documents are segmented into paragraphs, and each paragraph is represented by aggregated embeddings. Retrieval is then performed by computing the similarity between the query embedding and the embeddings of documents or their paragraphs, with the most relevant results being returned. Although effective for basic textual analysis, this approach is limited in its ability to capture contextual nuances and accommodate personalized representations.

\textbf{LLM-based Retrieval.} With the advent of LLMs, retrieval system architectures have evolved to harness their enhanced language understanding and feature integration capabilities. In the LLM-based retrieval architecture, traditional language models are replaced with an Llama-Tiny-based LLM to generate sophisticated embeddings for both queries and documents. In this scheme, query and document representations are generated by dedicated LLMs: the query LLM incorporates user-specific features alongside the query text, while the document LLM integrates both document content and document-specific attributes. This dual-model approach enables the system to capture personalized and contextual information within a unified embedding space.
A key advantage of the LLM-based retrieval approach is its seamless integration of features from multiple sources and modalities. The retrieval architecture is extended to support multi-modal inputs; for example, the document LLM is capable of processing not only textual information but also images and other non-textual data. This results in comprehensive, unified embeddings that more accurately represent the diversity of real-world content.

Overall, the shift toward LLM-based reference retrieval represents a significant milestone in the advancement of intelligent retrieval systems. By leveraging advanced language models and integrating multidimensional features, these systems achieve higher retrieval accuracy while delivering a more personalized and comprehensive user experience.

\subsubsection{Lightweight Ranking}
\begin{wrapfigure}{r}{0.5\textwidth}
    \centering
    \includegraphics[width=0.48\textwidth]{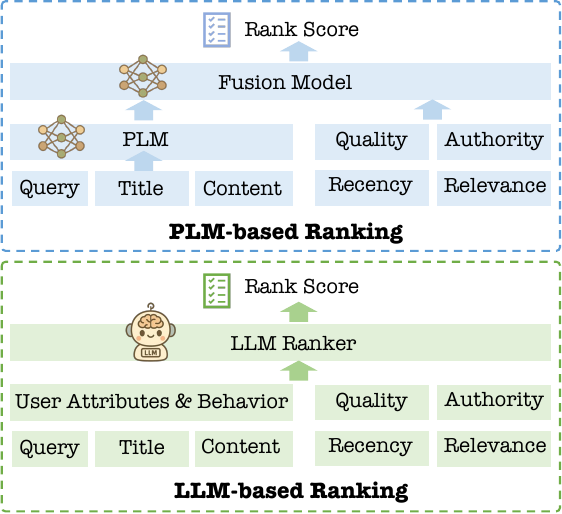}
    \caption{An illustration of the comparison of PLM-based Ranking and LLM-based Ranking.}
    \label{exe_ranking_compare}
    \vspace{-5pt}
\end{wrapfigure}

Building upon advancements in retrieval methods, the ranking component plays a crucial role in refining search results and ensuring that the most relevant documents are delivered to users. Traditionally, ranking strategies relied on PLMs to fuse multiple signals and produce a final relevance score. However, with the rapid development of LLMs, the ranking paradigm has experienced a significant transformation. Fig.~\ref{exe_ranking_compare} elucidates these changes, visually mapping the evolution from conventional PLM-based ranking to the LLM-based ranking.

\textbf{PLM-based Ranking.} In PLM-based ranking systems, traditional end-to-end rankers harness robust PLMs, such as those based on BERT or ERNIE~\citep{li2025s,li2025rankelectra,li2023s2phere,DBLP:journals/ml/LiXKBWCY24,DBLP:conf/icdm/0006XKSCWY23,DBLP:journals/tkde/LiXWKLLBWCDY23,DBLP:conf/pkdd/LiXKWSCCY23,li2025rankexpert}, to encode the primary features of each document, including the query, title, and content. Additional signals such as authority, quality, recency, and relevance features are fused through a neural network fusion layer to generate a final rank score~\citep{li2025m2oerank}. This end-to-end framework integrates both semantic and heuristic features, but it is limited by the representational power of the base model.

\textbf{LLM-based Ranking.} In the AI search paradigm, LLM-based rankers use embeddings generated by LLMs to represent the query, title, and content. These semantic embeddings are then combined, again using a neural network fusion layer, with external features such as authority, quality, recency, and relevance features. The use of LLM embeddings enables deeper contextual understanding and more sophisticated feature interactions, leading to more accurate ranking scores. Furthermore, a reasoning LLM is employed directly as the ranking model, processing all available features, including query, title, content, click behavior, recency, authority, quality, and user-specific attributes, in a unified manner to produce the final ranking score. This holistic, end-to-end approach enables effective reasoning over heterogeneous information sources, thereby significantly enhancing overall ranking performance.

The transition from traditional PLM-based rankers to LLM-based ranking architectures represents a paradigm shift in reference retrieval systems. By harnessing the deep semantic comprehension and sophisticated fusion capabilities of large language models, modern ranking systems achieve unprecedented improvements in accuracy and user satisfaction.


\subsection{LLM-Augmented Features}
Conventional feature engineering relies on handcrafted rules and shallow models to extract key signals, such as authority, recency and relevance. While this approach brought some level of semantic understanding, it suffered from several drawbacks:
\begin{itemize}
    \item \emph{Insufficient Coverage}. Limited ability to capture the full breadth of user intent and content diversity.
    \item \emph{Low Accuracy}. Limited semantic understanding led to suboptimal retrieval results.
    \item \emph{Difficult Maintenance}. Frequent manual updates and tuning were required as user needs and data evolved.
\end{itemize}

As LLMs advance in understanding language, context, and multi-modal data, their capability to generate rich semantic representations offers a promising avenue to enhance feature quality. Incorporating LLM-augmented features enables the system to more effectively capture latent user intentions, content quality, and contextual relevance, thereby enhancing retrieval precision. This integration also enables the system to process multimodal information, such as images and videos, thereby expanding the scope and robustness of relevance signals. Ultimately, leveraging LLM-generated features drives the development of more accurate, adaptable, and comprehensive retrieval systems. 

Specifically, the AI search system is enhanced with semantic features derived from LLMs. These LLMs directly process both the user query and the document text, yielding richer textual representations and facilitating a more nuanced assessment of factors such as authority, quality, and recency. 
This approach substantially improves the coverage and accuracy of the retrieval process, as LLMs can infer latent user needs and contextual relevance from the input data. Furthermore, the integration of LLMs with visual language models (VLMs) enables the system to process not only textual information but also multi-modal features, including videos and images.
Such multi-modal processing allows the system to capture a broader and more robust set of signals related to authority and quality, thereby enhancing retrieval relevance and user satisfaction. The shift from rule-based and model-based features to those augmented by LLMs and VLMs represents a transformative advancement in reference retrieval systems. By leveraging these powerful language and vision-language models, modern systems achieve broader coverage, higher accuracy, and increased adaptability.

For training with LLM-augmented features, the process begins by mining user session data to capture fine-grained behavioral patterns, including prior queries, user interactions, and content preferences. This comprehensive behavioral dataset, when combined with large-scale pretraining corpora, enables the system to develop a robust understanding of user intent and content relevance. During the training phase, the Search LLM is exposed to hundreds of billions of tokens via next-token prediction and self-supervised learning. This extensive exposure allows the LLM to acquire not only surface-level semantic knowledge but also deeper relational insights, such as authority, recency, and contextual quality, which are central to the LLM-augmented features described above.

By integrating LLM-augmented features into both the training and inference processes, the AI search system achieves significant advancement over traditional retrieval systems. The resulting system is capable of processing and synthesizing multi-modal signals, including text, images, and video, assessing authority and quality from multiple perspectives, and dynamically adapting to user-specific requirements. This comprehensive approach not only enhances the accuracy and robustness of the reference retrieval system but also ensures its scalability and maintainability in real-world, online environments.

\section{LLM-based Generation}
\subsection{Background}
In the first two stages, AI search paradigm focuses on the design and optimization of the task planner and task executor. Given the execution results and relevant documents retrieved by the Executor, it is critical to design a Writer agent within the AI search system to produce the final, accurate answer. However, the documents retrieved by the Executor frequently contain noise and errors, and addressing these issues to enhance the robustness of the model remains an urgent technical challenge.
Furthermore, to provide users with accurate and satisfactory answers, the AI search system is required to align its generated responses with users' preferences. Typically, users prefer that the information provided in the final answer conforms to the \textbf{Three H Standard}, \ie, \underline{H}elpfulness, \underline{H}armlessness, and \underline{H}onesty. 
Ensuring that the outputs of a RAG system align with user preferences and adhere to the three H standard remains a critical technical challenge that necessitates prompt resolution.
Ultimately, the AI search system generates and presents a final answer to the user, which in turn elicits a variety of feedback signals reflective of the user’s satisfaction. Explicit signals, such as subsequent query modifications, are readily observable; however, users also provide a wealth of implicit feedback through behaviors like click-through activity and dwell time on the presented results. Effectively leveraging both the explicit and implicit user feedback to refine the AI search system, particularly the Writer, remains a formidable technical challenge. 

Building upon the above motivation, we detail the design behind the Writer in the AI search system, focusing on the perspectives of \emph{Robustness}, \emph{RAG Task Alignment}, \emph{Optimization with User Feedback}, and \emph{Multi-Agent Joint Optimization}.

\subsection{Robust RAG System}
\begin{minipage}{\textwidth}
In practical RAG scenarios, the retriever is not always reliable and may retrieve noisy or 
\end{minipage}
\begin{wrapfigure}{r}{0.5\textwidth}
    \centering
    \includegraphics[width=0.5\textwidth]{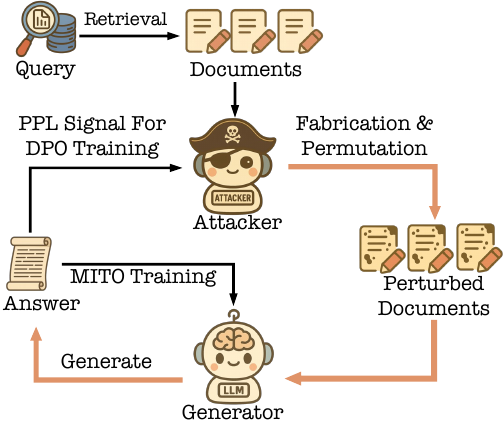}
    \caption{An illustration of ATM, which leverages a multi-agent system combined with adversarial tuning to enhance the robustness of the generator.}
    \label{rag_ATM}
    \vspace{-10pt}
\end{wrapfigure}
irrelevant documents. Consequently, these noisy references may misguide the generator, resulting in inaccurate or less robust outputs from the RAG system.
To address this issue, the AI search system proposes an adversarial tuning in a multi-agent method \emph{ATM}~\citep{ATM}, which leverages a multi-agent system combined with adversarial tuning to enhance the robustness of the generator. An overview of the ATM is illustrated in Fig.~\ref{rag_ATM}.

\subsubsection{Adversarial Setups for Robust RAG}
The ATM system consists of two components: an \emph{Attacker} and a \emph{Generator}. The Attacker aims to introduce perturbations to the documents retrieved by the retriever, while the Generator seeks to withstand these adversarial perturbations and accurately answer the questions.

Specifically, the Attacker perturbs the document by fabricating fake knowledge and hiding 
useful knowledge, which can be divided into the following two steps:
\begin{itemize}
    \item \textbf{Fabrication Generation.} Due to the RAG generator’s high sensitivity to noise, the Attacker imitates top-ranked documents by fabricating knowledge that is semantically related to the query but is either useless or misleading, and inserts this fabricated content into the document list.
    \item \textbf{List Permutation.} Because the RAG generator is highly sensitive to the relative position of useful knowledge within the document list, the Attacker could randomly shuffle the document list and duplicate some of its elements. This manipulation brings irrelevant documents to the forefront, effectively hiding the useful knowledge.
\end{itemize}

Given the question and the list of perturbed documents generated from the Attacker, the Generator takes them as inputs and aims to identify and utilize all useful documents while ignoring noisy ones, thereby maintaining robustness and generating correct answers.
The goal of the Generator can be formalized to maximize the objective as
\begin{equation}
    G(a\mid q,D^\prime) - \mathrm{dist}\left[G(a\mid q, D), G(a\mid q,D^\prime)\right], \label{eq:generator_open}
\end{equation}
where $a$ denotes the ground truth answer, and $q$ represents the question. $D$ and $D^\prime$ refer to the original retrieved documents and perturbed documents from the Attacker, respectively. $G(\cdot)$ signifies the probability that the Generator produces the answer, and $\mathrm{dist}\left[\cdot\right]$ quantifies the distance between the two probabilities. Maximizing Eq.~\ref{eq:generator_open} implies that the Generator should perform similarly regardless of whether the documents have been perturbed, while still producing the correct answer. In this way, the capability and robustness of the Generator could be enhanced.


\subsubsection{Multi-Agent Adversarial Training}
The training process for the ATM system comprises two phases: \emph{Initial Training} and \emph{Iterative Adversarial Tuning}. 

In the initial training phase, the Generator is trained to acquire fundamental RAG capabilities, providing a better starting point for optimization in the subsequent adversarial tuning. Specifically, the initial training involves four supervised fine-tuning (SFT) tasks: (1) answering the question using the originally retrieved document, (2) answering the question with only one ground-truth document, (3) answering without any document, and (4) extracting the ground-truth document from multiple retrieved documents.

During the iterative adversarial tuning phase, the Attacker and Generator are trained in an alternating manner. The Attacker progressively amplifies the intensity of its adversarial perturbations, while the Generator concurrently refines its generative capabilities to effectively counter these escalating attacks. The training details of the iterative adversarial tuning are introduced as follows.
\begin{itemize}
    \item For the Attacker, the objective is to synthesize adversarial documents that maximally disrupt the Generator's ability to produce correct answers. Specifically, a fabricated document $d^\prime$ is deemed adversarially effective if it induces high perplexity in the Generator when attempting to generate the correct answer. Formally, given a query $q$ and a fabricated document $d^\prime$, we quantify the Generator's degradation by computing the perplexity ($\mathrm{PPL}$) of generating the correct answer $a$ as
\begin{align}
    \label{eq:ppl}
    \mathrm{PPL}_G(a \mid q,\{d^\prime\}) = \exp\Big\{- \frac{1}{T_a}\sum_{t=1}^{T_a}\log P_G(a_t \mid a_{<t}; q, \{d^\prime\})\Big\}.
\end{align}

The Attacker is trained using Direct Preference Optimization (DPO)~\citep{rafailov2023direct} with the loss function defined as
\begin{align}
\mathcal{L}_\mathrm{DPO}(A_\theta&; A_\mathrm{ref}) = -\Big[\log \sigma\big(\beta \log \frac{A_\theta(d_{win}^\prime\mid q, D)}{A_\mathrm{ref}(d_{win}^\prime\mid q, D)} - \beta \log \frac{A_\theta(d_{lose}^\prime\mid q, D)}{A_\mathrm{ref}(d_{lose}^\prime\mid q, D)}\big)\Big],
\label{eq:dpo_loss}
\end{align}
where $A_\mathrm{ref}$ is the un-aligned Attacker (reference model), $A_\theta$ is the current Attacker to be optimized. $d_{win}^\prime$ and $d_{lose}^\prime$ represent a pair of fabrications generated by the Attacker, $win$ denotes the one with a higher $\mathrm{PPL}$-based reward, and vice versa.

\item The Generator aims to produce as many golden answers as possible by leveraging the input documents, while maintaining robustness even when noisy documents are injected. To accomplish this, ATM introduces the Multi-agent Iterative Tuning Optimization (MITO) loss to optimize the Generator as follows:
\begin{align}
\label{eq:mito_loss}
\mathcal{L}_\mathrm{MITO} = \mathcal{L}_\mathrm{SFT}(a \mid &q, D^\prime) + \alpha\mathcal{L}_\mathrm{KL};
\end{align}
\begin{equation}
    \label{eq:sft_loss}
    \mathcal{L}_\mathrm{SFT}(a \mid q,D^\prime) = - \sum_{t=1}^{T_a}\log P_G(a_t \mid a_{<t}; q, D^\prime);
\end{equation}

\begin{equation}
\begin{split}
\mathcal{L}_\mathrm{KL} = \sum_{t=1}^{T_a} \mathbb{D}_\mathrm{KL}[P_G(a_t \mid a_{<t}; q, D) \\
\Vert\ P_G(a_t \mid a_{<t}; q, D^\prime)].
\end{split}
\end{equation}
The inclusion of the SFT term enables the Generator to maintain output correctness when processing attacked documents. Additionally, $\mathcal{L}_\mathrm{KL}$ serves to reduce the discrepancy in answer generation probabilities between the given normal document list and the attacked document list. $\alpha$ is a pre-set hyper-parameter.
\end{itemize}


\subsection{RAG Task Alignment}
To better harness the impressive question-answering capabilities of LLMs within RAG scenarios, it is essential to further align LLMs with the specific requirements of RAG tasks.
To address this challenge, this work introduces a preference alignment technique for the RAG scenario (PA-RAG)~\citep{PA-RAG}, which achieves comprehensive alignment between LLMs and RAG requirements through multi-perspective preference optimization.

\subsubsection{RAG Task Requriements}
The requirements of RAG tasks can be summarized as follows: (1) \emph{Response informativeness}: the generator must effectively leverage valuable documents to ensure the completeness of the answer. (2) \emph{Response robustness}: the generator should be resilient to interference from noisy documents, maintaining the accuracy and coherence of the responses. (3) \emph{Citation quality}: the claims generated by the generator must be properly cited, ensuring that the answers are traceable to the retrieved documents.

\subsubsection{PA-RAG Optimization Objectives}
The optimization goal is to enable the generator to fully utilize valuable documents (Correctness) and accurately cite the references corresponding to claims (Citation quality).

Fully utilizing valuable documents means all answers contained within the documents should be included in the generator’s output. Formally, let $G$ represent the generator, $x$ denotes the input, $D=\{d_1,d_2,...,d_n\}$ represent the documents retrieved by the retriever, $A=\{a_1,a_2,...,a_n\}$ represent the short answers contained in the documents, and $y$ denote the response generated by the generator. The correct response by the generator can be expressed as:
\begin{equation}
\begin{aligned}
    y &= G(x, D), \\
    \text{s.t.} & \forall a_i \in A, C(y, a_i) = \text{True}
\end{aligned}
\end{equation}
where $C(y, a_i) = \text{True}$ implies that the answer $a_i$ is included in $y$.

Accurate citation means that each claim is fully supported by the cited documents, and irrelevant citations are avoided. Formally, the statement, claim, and citation could be denoted as $s$, $c$, and $t$, respectively. The correct citation of references corresponding to claims can be expressed as follows:
\begin{equation}
y=\{s_1,s_2,...,s_n\},
\end{equation}
\begin{equation}
s_i=\{\text{``claim'':}c_i, \text{``citation'':}t_i=[t_{i1},t_{i2},...,t_{in}]\},
\end{equation}
\begin{equation}
\forall s_i \in y, \phi(\text{concat}(t_i),c_i)=1
\label{func:4}
\end{equation}
\begin{equation}
\begin{aligned}
\forall s_i \in y, \forall t_{ij} \in t_i, \phi(t_{ij}, c_i) = 1 \lor \phi(\text{concat}(t_i \setminus t_{ij}), c_i) = 0
\label{func:5}
\end{aligned}
\end{equation}

\begin{minipage}{\textwidth}
Here, $\text{concat}(t_i)$ denotes the concatenation of all cited documents, and $\phi(t,c)=1$ indicates 
\end{minipage}
\begin{wrapfigure}{r}{0.48\textwidth}
    \centering
    \vspace{5pt}
    \includegraphics[width=0.48\textwidth]{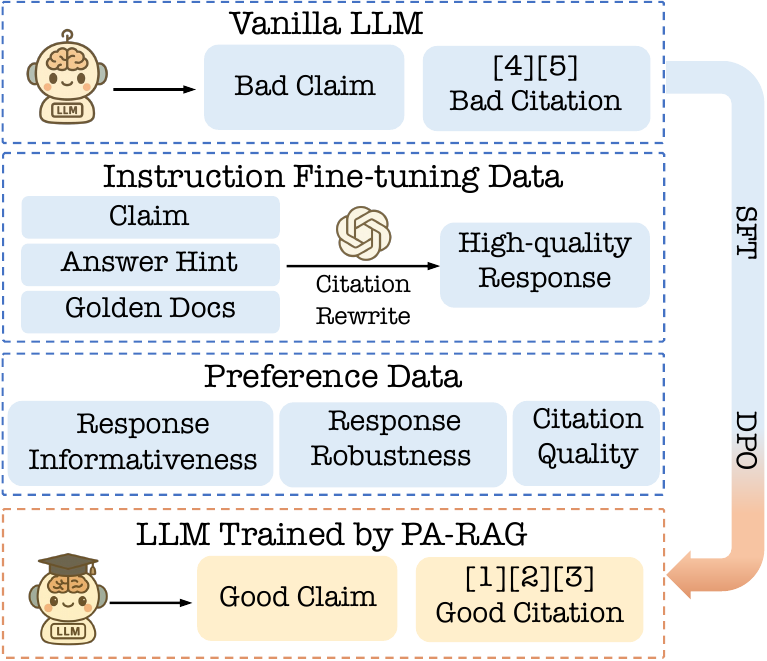}
    \caption{An illustration of PA-RAG, which achieves comprehensive alignment between LLMs and RAG requirements through multi-perspective preference optimization.}
    \label{rag_parag}
    \vspace{-40pt}
\end{wrapfigure}
that the claim $c$ is fully supported by the citation $t$.

\subsubsection{Methodology of PA-RAG}
The training of PA-RAG consists of two phases. The first phase involves instruction fine-tuning to equip the generator with fundamental RAG capabilities. The second phase employs multi-perspective preference optimization to further enhance the generator’s response informativeness, response robustness, and citation quality. An overview of the PA-RAG is illustrated in Fig.~\ref{rag_parag}.

\textbf{Instruction Fine-tuning for Basic RAG Capabilities. }
The instruction fine-tuning phase focuses on enabling the generator to equip the fundamental abilities to utilize and cite documents. The training data construction contains these three key steps as follows:
\begin{itemize}
\item \textbf{Construct questions and high-quality documents.}
Questions and corresponding short answers are sourced from existing datasets. 
To construct high-quality documents, the RAG retriever first retrieves the top 100 most relevant documents from the corpus for each question, then filters them to retain only those containing the short answers (referred to as ``golden documents'').
Subsequently, up to five documents are selected as prompt documents, ensuring coverage of all required answers and relevance to the query. 

\item \textbf{Construct high-quality responses.}
Responses are generated using GPT-4o, guided by prompts that include instructions, selected documents, and short answer hints. The detailed prompt can be illustrated as follows:
\label{appendix:prompt}
\begin{tcolorbox}[title=Prompt for GPT-4o]
\small
\textbf{Instruction:} Write an accurate, engaging, and concise answer for the given question using only the provided search results (some of which might be irrelevant) and cite them properly. Use an unbiased and journalistic tone. Always cite for any factual claim. When citing several search results, use [1][2][3]. Cite at least one document and at most three documents in each sentence. If multiple documents support the sentence, only cite a minimum sufficient subset of the documents. \\
\textbf{Qustion:} \{Question\}\\ 
The final answer should contain the following short answers: \{Short answers\} \\
\textbf{Documents:} \{Documents\} \\
\textbf{Answer:}
\end{tcolorbox}
\item \textbf{Citation rewrite.}
While GPT-4o produces coherent answers, direct outputs often exhibit inconsistent or incomplete citations, necessitating further refinement to meet quality standards. To address flawed citations, a three-step validation process is applied:

\emph{(1) Verification}: A Natural Language Inference (NLI) model evaluates whether cited documents logically support the associated claims.
\emph{(2) Correction}: If citations fail verification, traverse the power set of all prompt documents as citations to explore a feasible citation scheme.
\emph{(3) Simplification}: Redundant or irrelevant citations are pruned from valid subsets to retain only essential references.
\end{itemize}

\vspace{5pt}
\textbf{Preference Optimization of the Generator through DPO.} After the generator acquires the basic ability to utilize and cite documents, it further enhances response informativeness, response robustness, and citation quality sequentially through preference optimization. This phase requires creating preference data containing an input, a superior response (chosen output), and an inferior response (rejected output). 
\begin{itemize}
\item \textbf{Response Informativeness} refers to the completeness of the answer in the response. The optimization objective is to ensure the generator fully uses the document containing the short answer. The construction of preference data includes:
\emph{(1) Input Construction:} Like the input for instruction fine-tuning data, the input includes the instruction, the question, and high-quality prompt documents containing all short answers from up to 5 golden documents.
\emph{(2) Chosen Output Construction:} The chosen output is consistent with the output component of the instruction fine-tuning data, using GPT-4o and the \textit{citation rewrite} mechanism to construct a response that includes all short answers and the accurate citation.
\emph{(3) Rejected Output Construction:} To simulate cases where the generator misses parts of the golden documents, some golden documents are removed from the prompt, and generate inferior answers with the unoptimized generator.

\item \textbf{Response robustness} refers to the generator’s ability to resist interference. The optimization objective is to enable the generator to avoid disruption from noisy documents. The construction of preference data includes:
\emph{(1) Noisy document construction:} Noisy documents are divided into two types. The first type includes documents related to the question but lacking the answer; two such documents are randomly selected from the top 100 relevant retrieved documents without short answers. The second type consists of irrelevant documents to the question, with two randomly selected documents without short answers taken from retrieval results of other questions. \emph{(2) Input construction:} The input consists of the instruction, the question, and low-quality prompt documents combining up to five golden documents with four noisy documents. \emph{(3) Chosen output construction:} To simulate the generator ignoring all noisy documents, the chosen output is generated without including these documents, which is the same as the chosen output in the response informativeness part. 

\item \textbf{Citation quality} refers to the generator’s ability to accurately cite relevant documents. The optimization goal is to ensure correct citation of documents related to the claim while avoiding references to irrelevant sources.

Data construction involves two steps: first, generating responses and filtering to retain those containing all short answers; second, applying a \emph{citation rewrite} mechanism to identify incorrect citations that fail the NLI model verification or cite irrelevant documents as rejected output and then correct them as chosen output.

\end{itemize}
The preference optimization stage adheres to a strict optimization sequence, training the model sequentially and independently on response informativeness, response robustness, and citation quality. This sequence is arranged according to the increasing difficulty of preference optimization, similar to curriculum learning~\citep{bengio2009curriculum}. Modifying this order can result in performance degradation.

\subsection{Optimization with User Feedback}
Although LLMs demonstrate powerful capabilities in understanding human instructions and generating high-quality responses by pre-training on extensive corpora and subsequently fine-tuning on meticulously designed instruction-following datasets, it is often observed that fine-tuned LLMs produce unexpected and even harmful responses that significantly deviate from human expectations. To address this, most LLMs incorporate alignment strategies that aim to guide the models to meet predetermined dimensions, such as generating harmless, useful, and honest answers. Typically, these alignment methods~\citep{zheng2023secrets,touvron2023llama} first assign preference signals to answers generated by various models and then employ techniques, such as RLHF or DPO, to train the models based on these preference signals. 
However, these approaches usually incur high labor and time costs, as well as significant sampling inference overhead. Moreover, some studies~\citep{DBLP:journals/corr/abs-2306-11816,bai2022constitutional} have explored the use of AI-assisted annotation or AI-guided feedback as alignment signals to mitigate the reliance on human effort. Nevertheless, the feedback distributions from these sources generally do not align with online user behaviors and tend to be time-varying and heterogeneous. 
To tackle the above problems, the AI search system first explores the direct alignment of LLMs with online human behavior and proposes an LLM alignment method \underline{R}einforcement \underline{L}earning with \underline{H}uman \underline{B}ehaviors (\emph{RLHB}). Specifically, 
RLHB employs a multi-model simultaneous training mechanism with the target LLM as the generator and another auxiliary LLM as the discriminator. They are trained adversarially, ensuring that \textless query, response, feedback \textgreater~originates from real online interactions. In inference, the aligned generator accepts the user query along with the most preferred behavior as input signals to generate responses. 
Compared to RLHF, RLHB eliminates the need for extensive annotation and is adaptable to evolving user behaviors by updating behavior modeling in natural-language form.

The user behavior used in RLHB includes \emph{explicit} and \emph{implicit behavior}. More specifically, explicit behavior refers to users’ proactive feedback behavior, such as the click behavior on \emph{Like} and \emph{Dislike} buttons. The implicit behavior usually involves indirect and non-perceived interactions, such as \emph{Page View} (\emph{PV}) and \emph{Clicks} (\ie, the number of times an answer is clicked). In particular, \emph{Page View}, \emph{Clicks}, and \emph{Like} are positive indicators, and \emph{Dislike} is negative. 
At the implementation level, these indicators are first smoothed using a logarithmic transformation and then discretized into equal intervals. Subsequently, reward shaping is performed during reinforcement learning by computing $\left(\emph{Like} - \emph{Dislike}\right) / \left(\emph{PV} + \emph{Clicks}\right)$, integrating multiple reward signals into a holistic scalar. Next, we present the details of RLHB as follows.

The LLM alignment
with online human behaviors can be formulated as a Markov Decision Process (MDP) with state space $\mathcal{S}$, action space $\mathcal{A}$, transition dynamics $\mathcal{P}$, reward function $\mathcal{R}$ and discount factor $\gamma$, denoted as a tuple $(\mathcal{S}, \mathcal{A}, \mathcal{P}, \mathcal{R}, \gamma)$.
The AI search system considers human interactions and behaviors as the environment $E$. At each time step $t$, the Writer observes the current state $s_t \in \mathcal{S}$ from $E$ (the query triggered by the customer), and takes actions $a_t \in \mathcal{A}$ according to a policy $\pi: \mathcal{S} \mapsto p(\mathcal{A})$ that maps the states to a probability distribution over the actions.
After finishing the generation, the Writer receives a reward $r_t = \mathcal{R}(s_t, a_t)$ from the trained reward model and transitions to a new state $s_{t+1} \in \mathcal{S}$ determined by the environment $E$. The return of the interactive trajectory $\tau=\{s_1, a_1, \dots, s_T, a_T\}$ is defined as the cumulative $\gamma$-discounted rewards: $R(\tau) = \Sigma_{t=1}^{T}{\gamma^tr_t}$, where $T$ denotes the episode horizon. The objective is to optimize the policy $\pi$ by maximizing the expected returns from the initial state.

To guide the policy toward behavior that aligns with real-world demonstrations, RLHB incorporates a conditional discriminator $D(s_t, a_t; b_t)$ to determine whether a given state-action pair $(s_t, a_t)$, conditioned on feedback $b_t$, originates from real online environments. Here, $s_t$ denotes the state (query), $a_t$ denotes the action (response), and $b_t$ represents the behavioral feedback at time $t$. 
The discriminator is optimized by maximizing 
\begin{equation}
    \mathcal{L}_D = \mathbb{E}_{(s_t, a_t, b_t) \in \mathcal{M}_e}[\log(D(s_t, a_t; b_t))] + \mathbb{E}_{(s_t, a_t, b_t) \in \mathcal{M}_g}[1 - \log(D(s_t, a_t; b_t))],
\end{equation}
where $\mathcal{M}_e$ and $\mathcal{M}_g$ denote the expert and generated demonstration sets, respectively. 
The policy
$\pi(a_t | s_t, b_t)$ is 
optimized to generate feedback-conditioned responses using the clipped Proximal Policy Optimization (PPO)~\citep{schulman2017proximal} surrogate objective:
\begin{equation}
\mathcal{L}(\theta) = \mathbb{E}_t[\min(\ell_t \hat{A}_t, \text{clip}(\ell_t, 1-\epsilon, 1+\epsilon) \hat{A}_t)],
\end{equation}
where the importance ratio is defined as
\begin{equation}
\ell_t = \frac{\pi_{\theta}(a_t | s_t, b_t)}{\pi_{\theta_{\text{old}}}(a_t | s_t, b_t)}.
\end{equation}
Here,
$\theta$ represents
the current
policy parameters, and $\epsilon$ is the clipping
threshold. The advantage estimates $\hat{A}_t$ are computed using Generalized Advantage Estimation:
\begin{equation}
\hat{A}_t = \sum_{l=0}^{\infty}(\gamma \lambda)^l \delta_{t+l},
\end{equation}
with the temporal difference error
\begin{equation}
\delta_t = r_t + \gamma V_{\phi}(s_{t+1}; b_{t+1}) - V_{\phi}(s_t; b_t),
\end{equation}
where $\gamma$ is the discount factor, $\lambda$ is the GAE parameter, $V_{\phi}$ is the value function with parameters $\phi$
. The reward signal $r_t$ includes the discriminator output, regularized with a KL divergence penalty as
\begin{equation}
r_t - \eta \text{KL}(\pi_{\theta_{\text{old}}}(a_t | s_t, b_t), \pi_{\theta}(a_t | s_t, b_t)),
\end{equation}
with $\eta$ as a scaling coefficient. 
The value function $V_{\phi}$ is updated by minimizing the squared error:
\begin{equation}
\mathcal{L}(\phi) = \mathbb{E}_t[\|V_{\phi}(s_t; b_t) - \hat{R}_t\|^2],
\end{equation}
where $\hat{R}_t = \hat{A}_t + V_{\phi}(s_t; b_t)$ represents expected returns.
To improve training stability, a fraction $\kappa$ of high-reward demonstrations is bootstrapped as synthetic negative samples. Additionally, instead of numerical rewards, behavioral feedback is directly introduced as natural language instructions, leveraging the LLM’s capability to understand and respond to textual guidance.

\begin{figure*}[t]
    \centering
    \includegraphics[width=0.95\textwidth]{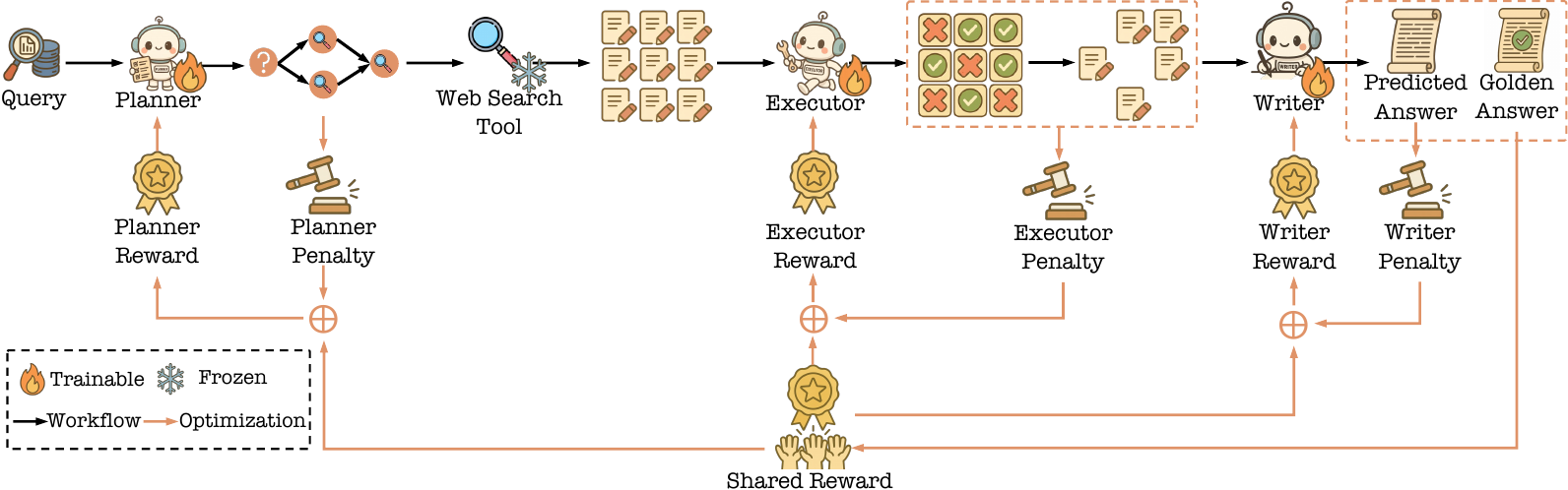}
    \caption{The illustration of MMOA-RAG, which leverages the Multi-Agent PPO algorithm to align the individual objectives of all modules with the shared goal of maximizing the quality and accuracy of generated answers.}
    \label{MMOA-RAG}
\end{figure*}

\subsection{Multi-Agent Joint Optimization}
LLMs have revolutionized various applications, including question answering~\citep{asai2023self}, information retrieval~\citep{xiong2024search}, and various forms of reasoning~\citep{hao2023reasoning}; however, their reliance on static pre-training inhibits timely updates to internal knowledge, often resulting in outdated or fabricated outputs~\citep{zhao2023survey}. RAG addresses this limitation by integrating current external knowledge into the generation process, thereby mitigating hallucinations and enhancing response accuracy. Contemporary RAG systems are structured as complex pipelines composed of interdependent modules, such as query rewriting~\citep{jiang2024rag}, first-stage retrieval~\citep{lewis2020retrieval}, re-ranking~\citep{salemi2024towards}, document processing~\citep{ke2024bridging}, and answer generation~\citep{shao2023enhancing}, that are traditionally optimized independently via supervised fine-tuning (SFT) on human-annotated data. This modular approach can lead to a misalignment between individual module objectives (e.g., training retrieval modules to maximize metrics like NDCG~\citep{dcg}) and the overarching goal of generating accurate answers, since improved relevance does not necessarily yield higher answer quality~\citep{cuconasu2024power}. 
Prior efforts~\citep{guu2020retrieval,lee2019latent,cuconasu2024power,izacard2020distilling,zamani2024stochastic,salemi2024learning, ke2024bridging} to overcome these challenges have employed end-to-end optimization techniques, including the propagation of rewards from the final answer through attention distributions or generation probabilities, as well as reinforcement learning methods like DPO and PPO; however, these strategies are typically confined to simplified two-module pipelines and fail to capture the intricate interdependencies and cooperative dynamics present in more sophisticated multi-module systems

To address the above limitations, the AI search system proposes a multi-module joint optimization algorithm for RAG (\emph{MMOA-RAG})~\citep{chen2025improving}. By modeling each intermediate module, such as the Planner, \emph{Web Search Tool}, the Executor, and the Writer, as an independent agent, the optimization process is formulated as a multi-agent collaborative RL task. MMOA-RAG leverages the multi-agent PPO (MAPPO) algorithm~\citep{yu2022surprising} to align the individual objectives of all modules with the shared goal of maximizing the quality and accuracy of generated answers, as evaluated through metrics like the F1 score against ground-truth responses. 
MMOA-RAG offers notable flexibility in its choice of reward functions and pipeline configurations, allowing it to accommodate diverse system designs while maintaining the cooperative alignment required for efficient optimization. Unlike earlier approaches, which often focus on optimizing simplified one-module or two-module RAG systems or fail to account for the interdependencies among components, MMOA-RAG provides a robust and generalizable framework for optimizing complex, multi-module RAG pipelines. In contrast to methods based on DPO or single-agent PPO, MMOA-RAG excels at fostering effective collaboration among modules, ensuring that the entire system operates cohesively to produce high-quality responses to knowledge-intensive tasks. Next, we will detail the proposed framework through the following algorithmic description.

As shown in Figure~\ref{MMOA-RAG}, MMOA-RAG consists of four components. Specifically, the Planner refines the initial query $q$, breaking down complex or ambiguous questions into manageable sub-questions $subq$ to improve retrieval efficiency. \emph{Web Search Tool} then gathers relevant documents for each sub-question, producing a candidate document set $D$. Given the candidate document set, the Executor contained in the Executor then identifies a refined subset $D_{\text{selected}}$ that most effectively supports generating the final answer. Using this filtered subset, the Writer produces the predicted response $Ans_{\text{predict}}$ to the original query. The Planner, Executor, and Writer are implemented using LLMs, enabling them to operate as RL agents that update their parameters based on reward signals. For efficiency, these modules share a single LLM architecture. As \emph{Web Search Tool} is not modeled as an RL agent due to its unique requirements, it is treated as part of the external environment in reinforcement learning.

The core focus of the MMOA-RAG framework lies in the collaborative optimization of the multiple modules in the pipeline. Its objective is to align the individual optimization goals of each component with the overarching task of producing high-quality responses. To this end, the framework uses a shared reward, $R_{\text{shared}}$, which measures the system's overall performance. A common choice for $R_{\text{shared}}$ is the F1 score of the Generator’s predicted answer, $Ans_{\text{predict}}$, calculated against the ground-truth response.

Given the inherently cooperative nature of the modules in the RAG pipeline, this shared reward, $R_{\text{shared}}$, serves as a unified signal to guide the optimization of all RL agents. This approach builds on established practices in multi-agent reinforcement learning (MARL) literature \citep{yu2022surprising, rashid2020monotonic, chen2022ptde}, where shared rewards are leveraged to promote collaboration among agents.
To ensure training stability and enhance convergence during the joint optimization process, the framework incorporates penalty mechanisms for individual agents. These penalties refine agent behaviors and discourage inefficient actions described as follows:
\begin{itemize}
    \item $P_{\text{QR}}$ refers to the penalty term for the Planner, designed to prevent the generation of an excessive number of sub-questions.
    \item $P_{\text{S}}$ refers to the penalty term for the Executor, used to avoid redundant or invalid document selections.
    \item $P_{\text{G}}$ refers to the penalty term for the Writer, discouraging the production of lengthy or verbose answers.
\end{itemize}
The interplay of the shared reward $R_{\text{shared}}$ and these penalty terms $P_{\text{QR}}$, $P_{\text{S}}$, and $P_{\text{G}}$ ensures coordinated optimization of all modules, promoting synergy and effectiveness across the pipeline.

\textbf{Detailed Configuration.} 
MMOA-RAG models three key modules, the Planner (QR), the Executor (S), and the Writer (G), as RL agents. These agents, defined within the tuple $\langle \mathcal{G}, O, A, R \rangle$, operate collaboratively to maximize the shared reward $R_{\text{shared}}$. $\mathcal{G} = \{\text{QR, S, G}\}$ is the set of agents. For each agent $i \in \mathcal{G}$, $O_i$ denotes its observation, $A_i$ represents its action space, and $R_i$ is its reward.

Concretely, for the Planner, the observation space $O_{\text{QR}} = \{Prompt_{\text{QR}}, q\}$ comprises a guiding prompt $Prompt_{\text{QR}}$ and the initial query $q$, while the action space $A_{\text{QR}}$ corresponds to the LLM's vocabulary space. The reward function $R_{\text{QR}}$ combines the shared reward $R_{\text{shared}}$ with a penalty term $P_{\text{QR}}$, designed to discourage generating an excessive number of sub-questions. More specifically, $P_{\text{QR}} = -0.5$ when the number of sub-questions exceeds 4, and is 0 otherwise. The reward of $QR$ can be represented as follows:
\begin{align}
    R_{\text{QR}} = R_{\text{shared}} + P_{\text{QR}},
\end{align}

Moreover, the observation of the Executor can be represented as $O_{\text{S}} = \{Prompts, q, D\}$, where $Prompts$ are guiding instructions, $q$ is the query, and $D$ is a set of $K$ candidate documents. Its action space $A_{\text{S}} = \{\text{"0"}, \text{"1"}, ..., \text{"K-1"}, \text{"Document"}, \text{","}\}$ is constrained to document selection, reducing the exploration space for efficient training. The reward function $R_{\text{S}}$ incorporates the penalty $P_{\text{S}}$, which is set to $-1$ when the selected document IDs are invalid (e.g., duplicates or incorrect formats) and 0 otherwise. $R_{\text{S}}$ can be represented as follows:
\begin{align}
    R_{\text{S}} = R_{\text{shared}} + P_{\text{S}},
\end{align}

Similarly, the observation of the Writer can be represented as $O_{\text{G}} = \{\text{Prompt}_{\text{G}}, q, D_{\text{selected}}\}$, where $Prompt_{\text{G}}$ guides the generation process, $q$ is the query, and $D_{\text{selected}}$ is the filtered set of documents provided by the Executor. The Writer shares the same action space as the Planner, $A_{\text{G}} = A_{\text{QR}}$. Its reward function can be formulated as 
\begin{align}
    R_{\text{G}} = R_{\text{shared}} + P_{\text{G}},
\end{align}
where $R_{\text{G}}$ introduces a penalty term $P_{\text{G}}$, which discourages excessively verbose answers by setting $P_{\text{G}} = -0.5$ when the generated content exceeds a predefined length, and 0 otherwise. Overall, these components ensure that all agents are aligned with the overarching goal of generating high-quality answers while maintaining computational efficiency and training stability.

\textbf{Warm Start with SFT.}
To facilitate joint optimization across multiple modules in MMOA-RAG, it is essential to perform SFT as a warm start for the trainable modules: the Planner, Executor, and Writer. The purpose of this warm start is to improve the modules' adherence to task-specific instructions, reduce the exploration space during MARL, and enhance the efficiency of exploration and exploitation during training. 

Specifically, for the Planner, public datasets from Rewrite-Retrieve-Read pipelines~\citep{ma2023query} are utilized for SFT by training the module to reformulate an input query $q$ into a series of sub-questions $\text{subq} = \{q_{\text{subq}_1}, q_{\text{subq}_2}, \dots\}$, making retrieval more effective. 

As for the Executor, its task is to identify a subset of candidate documents $D_{\text{selected}}$ from a given pool $D = \{d_1, d_2, \dots, d_K\}$ that are most relevant to answering $q$. To construct the Executor’s SFT dataset, a heuristic approach is employed. First, the stop words and punctuation are removed from $q_i$, retaining only its essential keywords to form a filtered query set $\text{Set}_{q_i}$. These keywords are then compared against the content of each candidate document $d_{i,j}$ of $q_i$ to compute a subset $\text{Set}_{d_{i,j}}$. The document $d_{i,j}$ is included in $D_{\text{selected}}$ if $\text{Set}_{q_i} \cap \text{Set}_{d_{i,j}} \neq \emptyset$, providing the ground-truth labels for training. 

Meanwhile, the Writer is fine-tuned to synthesize the final answer $\text{Ans}_{\text{predict}}$ based on $D_{\text{selected}}$ and the query $q$, using $\text{Ans}_{\text{golden}}$ as the ground-truth output. The supervised loss for the Writer, as well as for other modules, can be formulated as
\begin{align}
    L_{\text{SFT}}(\theta) = -\frac{1}{N} \sum_{n=1}^{N} \log P(Y_i \mid X_i; \theta),
\end{align}
where $N$ is the number of training samples, $X_i$ represents the input (e.g., $q$ and $D_{\text{selected}}$), $Y_i$ is the ground-truth output (e.g., $\text{Ans}_{\text{golden}}$), and $\theta$ are the model parameters. By leveraging these carefully constructed SFT datasets for each module, MMOA-RAG ensures effective initialization, enabling the modules to work in synergy during joint multi-agent optimization.

\textbf{Multi-Agent Optimization.}
Following the warming start with SFT, the LLM is better equipped to follow precise task-specific instructions, enabling improved performance across three agents. To further enhance system performance, MMOA-RAG employs a fully cooperative MARL strategy based on MAPPO. In this setup, the modules collectively optimize their policies under a shared global reward, $R_{\text{shared}}$, reflecting the overall quality of answers produced by the system (e.g., F1 score of $\text{Ans}_{\text{predict}}$). The modules, the Planner (QR), the Executor (S), and the Writer (G), share parameters through a common LLM backbone, effectively reducing computational overhead while promoting efficient collaboration between agents. The optimization involves three key models: the Actor model (with parameters $\theta$), which generates outputs like $\text{subq}$, $D_{\text{selected}}$, and $\text{Ans}_{\text{predict}}$ based on observations $O_i$; the Critic model (with parameters $\phi$), which estimates the state-value function $V_\phi$ for stability in optimization; and the pre-trained SFT model (with parameters $\theta_{\text{SFT}}$), which serves as the baseline reference. The overall loss function integrates the Actor and Critic losses as
\begin{align}
    \mathcal{L}(\theta, \phi) = \mathcal{L}_{\text{Actor}}(\theta) + \alpha \cdot \mathcal{L}_{\text{Critic}}(\phi),
\end{align}
where $\alpha$ is a hyperparameter balancing the two terms. The Actor loss $\mathcal{L}_{\text{Actor}}$ incorporates advantage estimates $\hat{A}_{t}^i$, which is computed using Generalized Advantage Estimation (GAE)~\citep{schulman2015high}, and the importance sampling ratio $r_t^i$ to perform stable policy updates:
\begin{align}
    \mathcal{L}_{\text{Actor}}(\theta) = \sum_{i} \sum_{t} \min \big[ r_t^i \hat{A}_{\pi_{\theta}}^{i,t}, \text{clip}(r_t^i, 1-\epsilon, 1+\epsilon) \hat{A}_{\pi_{\theta}}^{i,t} \big],
\end{align}
where $r_t^i = \pi_\theta(a_t^i \mid s_t^i)/\pi_{\theta_{\text{old}}}(a_t^i \mid s_t^i)$. Meanwhile, the Critic loss $\mathcal{L}_{\text{Critic}}$ minimizes the state-value prediction error as:

{\footnotesize
\begin{align}
    \mathcal{L}_{\text{Critic}}(\phi) = \sum_i \sum_t \max \left[ (\Delta V_{i,t})^2, \left( \text{clip}\left(V_{\phi}^{i,t}, V_{\phi_{\text{old}}}^{i,t} \pm \epsilon\right) - V_{\text{target}}^{i,t} \right)^2 \right],
\end{align}
}

Here, $\Delta V_{i,t} = V_{\phi}^{i,t} - V_{\text{target}}^{i,t}$, where $V_{\phi}^{i,t} = V_{\phi}(s_t^i)$. The term $V_{\text{target}}^{i,t}$ represents the cumulative return and $s_t^i$ is the state-value function.

The reward function $R(s_t^i, a_t^i)$ combines $R_{\text{shared}}$, agent-specific penalties $P_i$, and a regularization term penalizing deviations from the SFT-informed policy as:
\begin{align}
    R(s_t^i, a_t^i) =
    \begin{cases} 
        R_{\text{shared}} + P_i - \beta \log \frac{\pi_\theta(\text{Answer}_i \mid O_i)}{\pi_{\theta_{\text{SFT}}}(\text{Answer}_i \mid O_i)}, & \text{if } t = T, \\
        0, & \text{otherwise}.
    \end{cases}
\end{align}
The entire training process involves collecting rollouts for each agent, estimating advantages with GAE, and updating the Actor and Critic parameters by optimizing $\mathcal{L}(\theta, \phi)$. Additionally, to speed up the entire training process, we can perform a minibatch in the rollout process. Finally, we can obtain a well-trained Actor model used for the next inference and evaluation.

\section{Light-Weighting LLM}
\label{sec:lightweighting_llm_generation}

\subsection{Background}
    \label{ssec:motivation}

    \begin{figure*}
        \centering
        \includegraphics[width=0.93\textwidth]{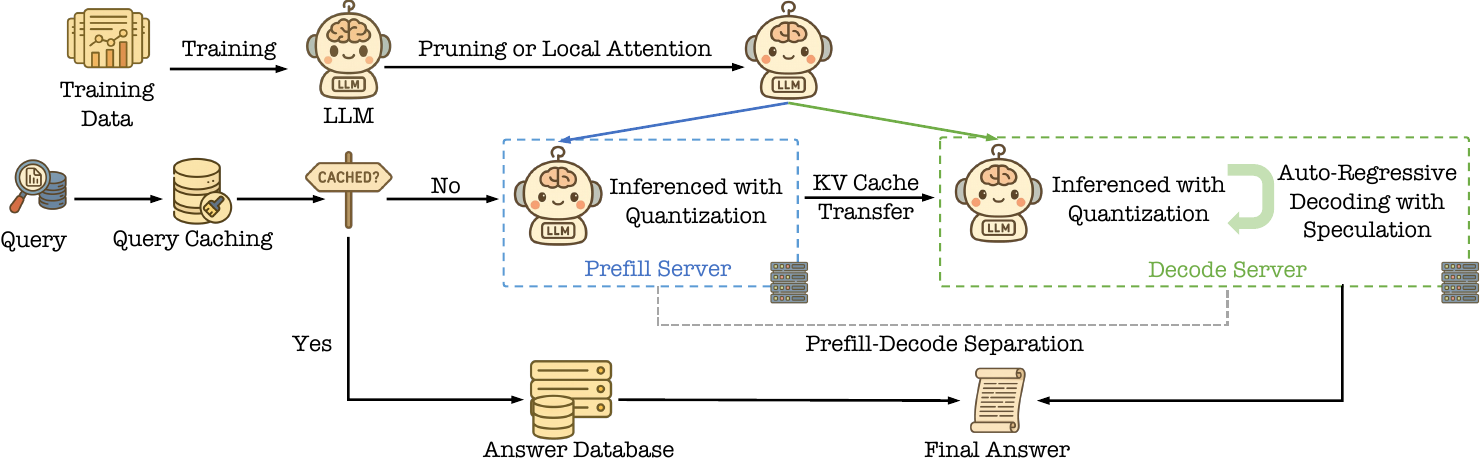}
        \caption{\centering The technical pipeline of lightning LLM’s generation.}
        \label{fig:lightning:overview}
    \end{figure*}

    AI-powered search applications have experienced a substantial increase in user traffic. For example, prominent platforms such as Perplexity are reported to manage over 10 million daily visits~\citep{perplexityai}. The core technology driving these applications consists of LLMs, which predominantly utilize the Transformer architecture~\citep{DBLP:conf/nips/VaswaniSPUJGKP17}. Although the Transformer model has demonstrated notable effectiveness, its computational complexity, which scales quadratically with the input sequence length, introduces significant challenges during inference. Moreover, in adherence to scaling laws~\citep{scalinglaw:arxiv20:Kaplan}, LLMs have evolved to encompass an exceptionally large number of parameters. For instance, models such as LLaMA-4, DeepSeek-V3, and Qwen-3 possess parameter counts ranging from 100 billion to 1,000 billion, thereby requiring considerable storage capacity.

    The computational requirements of LLMs necessitate considerable hardware resources, and their inference latency constitutes a significant portion of the overall response time in AI search applications. For example, a state-of-the-art LLM for AI search, DeepSeek-R1~\citep{deepseekai2025deepseekr1incentivizingreasoningcapability}, which reportedly incurs high deployment costs (requiring 32 GPUs, each with 80 GiB memory per instance) and suffers from latency bottlenecks (a first token latency of 2.91 seconds and a decoding speed of approximately 30 tokens/second). Search applications are inherently latency-sensitive, as users typically have limited tolerance for waiting times.


    Consequently, optimizing LLM inference has emerged as a critical imperative to reduce latency, minimize operational costs, and enhance overall system efficiency. This section examines lightweighting methodologies that have gained widespread adoption in both industrial and academic settings due to their demonstrated effectiveness, feasibility of deployment, cost efficiency, and ability to deliver measurable performance improvements beyond theoretical gains. Lightweighting techniques for LLMs in AI search can be systematically categorized into two principal approaches: (1) algorithmic-level optimizations, which focus on reducing model parameter counts and computational complexity, and (2) infrastructure-level optimizations, which enhance inference efficiency through improved computational workflows and resource utilization strategies.

\subsection{Algorithmic-Level Lightweighting}
\label{ssec:algorithmic_lightweighting}
Algorithmic-level lightweighting primarily targets the reduction of an LLM's parameter count and inherent computational complexity. Reducing the number of parameters directly lessens storage requirements, while decreasing computational load alleviates the demand on computing resources. In the following paragraphs, we introduce two highly effective and low-deployment-cost techniques for reducing both computational and parameter load: local attention and pruning.

    
\subsubsection{Local Attention}
The internal computations of LLMs are predominantly characterized by matrix multiplications involving input context tokens and model parameter matrices. A prominent strategy to mitigate this computational burden is to restrict the scope of input tokens attended to by each token, a technique termed \textbf{local attention}. It is well-established that the computational complexity of the standard full attention mechanism exhibits quadratic scaling ($O(N^2)$) with respect to the input sequence length $N$. However, a substantial body of research~\citep{xiao2023efficient} suggests that full attention is not always necessary, as only a subset of contextual tokens typically exerts a significant influence on the model's output. Consequently, local attention mechanisms can often serve as an effective substitute for full attention.
    
While various lightweight attention mechanisms have been proposed, such as linear attention~\citep{minimax2025minimax01scalingfoundationmodels}, RWKV~\citep{peng2023rwkvreinventingrnnstransformer}, and Mamba~\citep{gu2024mambalineartimesequencemodeling}, their adoption typically requires training models \textit{de novo}, incurring substantial costs. In contrast, local attention techniques can often be integrated into pre-trained full-attention models with minimal overhead. This adaptation can be accomplished either through lightweight fine-tuning~\citep{zhang2025lighttransfer} or, in some cases, applied directly at inference time without any retraining, as demonstrated by methods like StreamingLLM~\citep{xiao2023efficient}.

Recent works have explored various strategies to adapt pre-trained models. For instance, LightTransfer~\citep{zhang2025lighttransfer} addresses the high computational cost of full attention by identifying and modifying ``lazy layers'' where attention patterns are highly localized. Its innovation lies in selectively replacing full attention in these layers with a more efficient mechanism like streaming attention. This approach significantly reduces the KV cache footprint and computational load while preserving long-context capabilities, and can be applied either training-free or with minimal fine-tuning.

More advanced approaches aim to linearize pre-trained Transformers to achieve linear-time inference. A key challenge in this conversion is integrating recurrent mechanisms, such as gates, without costly retraining. {Liger}~\citep{lan2025liger} tackles this problem by repurposing existing weights from the pre-trained model's key projection to construct these gating mechanisms, thereby avoiding new parameters. Its hybrid attention, combining local sliding window attention with a linear recurrent model, effectively balances local context processing with long-range dependency modeling.

Similarly, {LoLCATs}~\citep{zhang2025lolcats} focuses on converting standard softmax attention into a subquadratic alternative, with a strong emphasis on scalability to extremely large models. The core innovation is a two-stage framework: an ``Attention Transfer'' step first trains a subquadratic mechanism to mimic the original attention, followed by a ``Low-rank Linearizing'' step that uses LoRA to correct residual errors. This parameter-efficient method has successfully linearized models as large as 405B parameters, achieving up to a 3$\times$ throughput increase while maintaining performance comparable to the original Transformer.

Furthermore, models adapted with local attention can leverage existing, highly optimized attention libraries, such as FlashAttention~\citep{dao2022flashattentionfastmemoryefficientexact}, during deployment, thereby ensuring computational efficiency in practice.

\subsubsection{Model Pruning}
\textbf{Pruning} offers another powerful strategy for model lightweighting by directly removing redundant parameters from a model's weight matrices. This process reduces both the model's storage footprint and its computational demands during inference. Pruning techniques can be broadly categorized into unstructured, semi-structured, and structured pruning. While unstructured pruning removes individual weights, leading to sparse models that often require specialized hardware or libraries for acceleration, \textit{structured pruning} removes entire structural components, such as attention heads, neuron columns, or even entire layers. This results in a smaller, dense model that can be directly accelerated on standard hardware, making it highly practical for deployment. Similar to local attention, many advanced structured pruning methods can be applied to pre-trained LLMs with minimal fine-tuning or even in a training-free manner, offering a cost-effective path to model compression.

Recent works have developed various innovative approaches to structured pruning. For instance, {Layer Collapse (LaCo)}~\citep{yang2024lacolargelanguagemodel} introduces a training-free method that directly removes entire layers from a pre-trained model. Its core innovation is the \textit{Reserving-Differences-while-Seeking-Common (RDSC) Layer Merge} principle, which calculates the parameter differences between adjacent layers and merges this differential into a subsequent layer. This ``collapses'' layers, effectively pruning them while preserving the model's overall function. A key advantage of LaCo is its ability to remove 30\% to 50\% of layers while maintaining performance and, crucially, preserving the model's internal architecture. This allows the pruned model to serve as a drop-in replacement in existing systems without code modifications.

Other methods focus on more granular structures. The CoFi ({Co}arse-to-{Fi}ne) pruning method~\citep{xia2022structuredpruninglearnscompact} demonstrates that structured pruning can achieve compression and speedup comparable to distillation but at a much lower computational cost. CoFi's innovation lies in simultaneously pruning both coarse-grained structures (e.g., layers) and fine-grained units (e.g., attention heads, hidden dimensions) using masks of varying granularity. To maintain accuracy, it integrates a dynamic layerwise distillation method that learns the optimal layer mapping between the teacher and the pruned student model, achieving over 95\% sparsity and 10$\times$ speedups while retaining high accuracy.

Building on classic compression frameworks, {SlimGPT}~\citep{ling2024slimgptlayerwisestructuredpruning} adapts the Optimal Brain Surgeon (OBS) technique for the unique challenges of structured pruning in LLMs. To overcome the limitations of OBS, SlimGPT introduces a \textit{Batched Greedy Pruning} algorithm that enables low-cost and rapid structured pruning by processing groups of parameters. It employs a grouped Cholesky decomposition for attention heads and a dynamic group size for FFNs to achieve near-optimal pruning efficiently. A key advantage of SlimGPT is its accessibility: it is a task-agnostic method requiring only a small calibration dataset and can complete the entire pruning process in about an hour on a single GPU.

Another line of work targets semi-structured pruning, which is critical for hardware acceleration but can often degrade model accuracy. A novel plug-and-play post-training pruning (PTP) method~\citep{zhang2024plug} addresses this by introducing two key components. First, a new metric, {Relative Importance and Activation (RIA)}, jointly considers weight magnitude and activation data to prevent the inadvertent removal of important weight channels, a drawback of previous metrics. Second, it employs an efficient {channel permutation} algorithm that reorganizes input channels to maximally preserve important weights before enforcing the N:M sparsity pattern. This training-free approach is scalable to models over 70B parameters and makes sparse LLM inference more practical by converting models to a hardware-friendly N:M format with minimal performance loss.

\subsection{Infrastructure-Level Lightweighting}\label{ssec:infrastructure_lightweighting}
Infrastructure-level lightweighting encompasses a diverse range of techniques aimed at optimizing the computational and memory efficiency of large-scale AI systems, particularly LLMs. These methods can be broadly categorized based on their applicability: some exploit the unique characteristics inherent in specific applications like AI search, while others offer general-purpose improvements to the underlying LLM inference pipeline. This section will delve into several key strategies. Specifically, we will first examine optimizations tailored to the distinct operational patterns of AI search applications, including those that leverage \textit{input-output similarity}, address \textit{output length reduction}, and capitalize on \textit{query semantic similarity}. Subsequently, we will discuss more general infrastructure enhancements applicable to a broader range of LLM use cases, such as \textit{prefill-decode separation}, \textit{quantization}, and \textit{speculative decoding}, which target fundamental aspects of the inference process.

\subsubsection{Output Length Reduction}\label{sssec:output_length_reduction}
A further distinguishing characteristic is the pronounced asymmetry in length between user input queries and LLM-generated outputs, where the latter are typically substantially longer. For instance, a concise query such as ``Explain the history of Hawaii'' may elicit an extensive response.
    
Numerous methods aim to mitigate the high inference costs associated with excessively long outputs. A primary strategy involves reducing the length of the model's generated responses. This can be achieved through several approaches:
\begin{itemize}
    \item \textit{Prompt-based Length Reduction.} Some techniques leverage prompting to curtail output length. For example, TALE-EP~\citep{han2024token} estimates token budgets by first prompting the LLM to determine a reasonable token limit for a given task, then incorporating this estimate into a constraint prompt that guides the model to generate concise yet accurate responses. Chain-of-Draft (CoD)~\citep{xu2025chain} addresses the verbosity common in step-by-step reasoning by instructing models to ``Think step by step, but only keep a minimum draft for each thinking step, with at most five words,'' thereby preserving essential intermediate reasoning while significantly reducing token usage. Lee et al.~\citep{lee2025llmscompresschainofthoughttoken} conducted a systematic study of different compression instructions, revealing a universal trade-off between reasoning length and accuracy, and proposing the concept of ``token complexity'' as the minimum number of tokens required for successful problem-solving.
    
    \item \textit{Training-based Length Reduction.} Other methods achieve output length reduction through modifications during the training phase. Demystifying~\citep{yeo2025demystifying} implements a Cosine Reward function based on Dirichlet principles, incorporating an ``exceed length penalty'' to control Chain-of-Thought (CoT) reasoning length while maintaining performance. Aggarwal et al.~\citep{aggarwal2025l1} modify the training data by incorporating explicit length constraint instructions, such as ``Think for N tokens,'' before applying policy optimization to pre-trained reasoning models. O1-Pruner~\citep{luo2025o1} introduces a Length-Harmonizing Reward that calculates the ratio between reference model and predicted CoT lengths, incorporating accuracy constraints to prevent performance degradation while shortening reasoning processes. DAST~\citep{shen2025dast} employs SimPO to fine-tune models using a specially constructed length-preference dataset based on a self-defined token-length budget measurement. Kimi~\citep{team2025kimi} incorporates a length penalty into its policy optimization approach to improve long CoT activations and facilitate model merging.
    
    \item \textit{Compression of Intermediate States.} A distinct category of techniques focuses on compressing intermediate data generated during the model's inference process. For instance, Coconut (Chain of Continuous Thought)~\citep{hao2024training} treats final-layer hidden states as ``continuous thought'' to replace discrete tokens, reusing these hidden states as subsequent input embeddings and gradually adding latent CoT tokens during training, thereby improving accuracy while reducing intermediate ``thinking'' tokens. CODI~\citep{shen2025codi} employs self-distillation where the model serves as both teacher and student, jointly learning explicit and implicit CoT while aligning hidden activations, enabling internal reasoning without generating explicit tokens. CCOT~\citep{cheng2024compressed} condenses long reasoning into short, contentful contemplation tokens by precomputing full CoT, selecting important hidden states, and training LoRA modules to predict and decode these key tokens.
\end{itemize}

\subsubsection{Semantic Caching}\label{sssec:query_semantic_similarity}
AI search applications frequently encounter a high degree of semantic similarity among queries submitted by different users. Consequently, a considerable fraction of distinct input queries can yield identical or substantially similar answers (e.g., ``Tell me about the history of Hawaii'' versus ``What is the history of Hawaii?''). This redundancy is leveraged by various optimization strategies, including query caching, result caching, and query rewriting~\citep{gill2025meancacheusercentricsemanticcaching}.
    
For instance, MeanCache~\citep{gill2025meancacheusercentricsemanticcaching} exemplifies a user-side semantic caching system tailored for LLM-powered web services. When a query is submitted to a MeanCache-enabled LLM service, the system computes an embedding vector for the query. This embedding is then compared against cached query embeddings using cosine similarity to identify potential matches. For each semantically similar query identified in the cache, MeanCache analyzes its associated context chain, comparing it with the dialogue history of the current query. If a match is found—that is, a cached query exhibits both semantic similarity and a congruent context chain—the corresponding response is retrieved from the local cache and served to the user. Otherwise, MeanCache forwards the query to the LLM service to elicit a response; this response, along with the query and its embedding, is subsequently stored in the cache. To preserve user privacy, MeanCache employs Federated Learning (FL) to train a compact embedding model, thereby obviating the need to store user data on a central server. The embedding model is trained using a multi-task learning approach that incorporates both contrastive loss and multiple-negatives ranking loss. This process also involves determining an optimal cosine similarity threshold, denoted as $\tau$, to strike a balance between true positives (cache hits) and false positives. To mitigate storage and computational overheads, MeanCache applies Principal Component Analysis (PCA) for dimensionality reduction of the embedding vectors. Empirical studies indicate that approximately 31\% of user queries exhibit similarity to previously submitted ones, underscoring the potential of user-side caching to significantly reduce LLM inference costs.

\subsubsection{Quantization}\label{sssec:quantization}
Quantization is a widely adopted infrastructure-level lightweighting technique. It involves reducing the numerical precision of the model's weights and/or activations (e.g., from 32-bit floating-point, FP32, to 8-bit floating-point, FP8, or even lower bit integers). This reduction in data representation bit-width decreases memory footprint, memory bandwidth requirements, and can accelerate computations on hardware supporting lower precision arithmetic. Existing quantization methods for Large Language Models (LLMs) can be divided into two categories: weight-only quantization and weight+activation quantization.

\begin{itemize}
    \item \emph{Weight-only Quantization.} In weight-only quantization, models like ZeroQuant-V2~\citep{yao2023zeroquantv2} apply uniform quantization to OPT and BLOOM, finding that 8-bit weights with 16-bit activations preserve performance while 4-bit weights cause significant degradation. They introduce Low-Rank Compensation (LoRC) to approximate quantization error with a storage-efficient low-rank matrix. GLM-130B~\citep{zeng2022glm} demonstrates successful 4-bit quantization using row-wise symmetric quantization, attributed to its well-shaped weight distributions. Non-uniform methods include LUT-GEMM/nuQmm~\citep{park2022nuqmm}, which extends binary-coding quantization with bias terms and group-wise quantization. SqueezeLLM~\citep{kim2023squeezellm} uses sensitivity-based k-means centroids for non-uniform quantization, optimizing for memory-bound LLM inference. QLoRA~\citep{dettmers2023qlora} introduces NormalFormat datatype based on Quantile Quantization, normalizing weights to standard normal distribution. GPTQ~\citep{frantar2023gptq} builds on Optimal Brain Quantization, quantizing rows in parallel with lazy batch updates and Cholesky reformulation for stability. QuIP~\citep{chee2023quip} defines optimal adaptive rounding methods using LDL decomposition and incoherence processing, achieving viable 2-bit quantization. AWQ~\citep{lin2023awq} preserves significant activation channels in FP16 while quantizing others, using activation-aware scaling to better represent salient weights.
    \item \emph{Weight+Activation Quantization.} For weight+activation quantization, LLM.int8~\citep{dettmers2022llm} identifies systematic outliers in activations, preserving outlier dimensions in high-precision while quantizing average values to INT8. RPTQ~\citep{yuan2023rptq} clusters and reorders activation dimensions based on min/max values, fusing operations to reduce latency. Low-bit floating-point formats like FP4/FP8 are explored in MoFQ~\citep{zhang2023integer} and ZeroQuant-FP~\citep{wu2023zeroquant}, providing better handling of activation outliers than integer formats. SmoothQuant~\citep{xiao2023smoothquant} suppresses outliers by scaling down activation dimensions and scaling up corresponding weight dimensions with a controllable hyperparameter. Outlier Suppression~\citep{wei2022outlier} uses inverse gamma values from LayerNorm as scaling factors, while Outlier Suppression+~\citep{wei2023outlier} adds shifting factors to remove asymmetry in activations. FPTQ~\citep{li2023fptq} introduces logarithmic activation equalization for non-linear moderation of activation distributions. OmniQuant~\citep{shao2023omniquant} learns optimal clipping ranges for extreme values, and QLLM~\citep{liu2023qllm} splits outlier channels into sub-channels while merging similar ones.
\end{itemize}

\subsubsection{Prefill-Decode Separating Deployment}\label{sssec:prefill_decode_separation}
The LLM inference process can be divided into two distinct phases: the \textit{prefill} phase, which processes the input prompt and is computationally intensive (compute-bound), and the \textit{decode} phase, which generates subsequent tokens autoregressively and is memory bandwidth-intensive (memory-bound). Recognizing these differing characteristics, mature LLM inference frameworks often implement separate deployment and resource allocation for these phases. For instance, systems like Mooncake~\citep{qin2024mooncakekvcachecentricdisaggregatedarchitecture}, or as implied by the architecture of models like DeepSeek-R1~\citep{deepseekai2025deepseekr1incentivizingreasoningcapability}, dynamically adjust the ratio of servers or resources dedicated to prefill and decode tasks based on user load and request patterns. This dynamic allocation ensures optimal utilization of computational resources, minimizing bottlenecks and improving throughput. For example, a system might allocate a certain number of nodes/GPUs for parallel prompt processing (prefill) and a different set for token generation (decode), scaling each pool independently.


    
\subsubsection{Speculative Decoding}\label{sssec:speculative_decoding}
Speculative decoding represents a significant advancement in accelerating large language model inference through parallel processing, and existing methods can be categorized into generation and refinement strategies. Generation approaches include predefined fill tokens, which use random initialization but require multiple refinement iterations; retrieval-based methods like REST~\citep{he2023rest} that leverage exact suffix matching from datastores; and N-gram techniques such as ANPD~\citep{ou2024lossless} that adapt predictions based on contextual patterns.
    
Auto-regressive generation represents the most common approach, split between independent and dependent drafters. Independent drafters like SpecDec~\citep{xia2023speculative} use smaller models to generate tokens verified by larger target models, with improvements like BiLD~\citep{kim2024speculative} adding confidence thresholds for verification. Dependent drafters reduce computation by sharing resources between draft and target models. Layer skipping methods like Draft\&Verify~\citep{zhang2023draft} selectively skip intermediate layers, while dependent head approaches like EAGLE~\citep{li2024eagle} add lightweight prediction heads to the target model's hidden states.
Multi-token prediction techniques generate multiple tokens simultaneously. Medusa~\citep{cai2024medusa} introduced parameter-efficient lightweight decoding heads on pre-trained models, while Amphista~\citep{li2024amphista} enhanced this with bi-directional self-attention to model inter-token relationships.
    
For refinement strategies, single-pass verification dominates the field. Linear verification methods sequentially validate draft tokens against the target model's distributions, with Fast Inference~\citep{leviathan2023fast} introducing speculative sampling to improve acceptance rates. Tree-based verification approaches like SpecInfer~\citep{miao2023specinfer} construct trees of possible completions for parallel exploration, with optimizations like Sequoia's~\citep{chen2024sequoia} hardware-aware tree optimizer maximizing computational efficiency. Iterative decoding methods permit multiple refinement iterations, with approaches like Jacobi~\citep{teng2024accelerating} reframing auto-regressive generation as an iterative optimization problem that can produce identical output to traditional decoding under greedy sampling.

\section{Evaluation}
\begin{table}[t]
\centering
\caption{Relative improvements of the proposed AI search system (\emph{AI Search}) to the legacy system (\emph{Web Search}) across query complex categories in the manual evaluation.}
\setlength\tabcolsep{2pt}
\renewcommand{\arraystretch}{1.3}
 {
 \small
 \begin{tabular}{l | P{0.23\linewidth}| P{0.32\linewidth} | P{0.23\linewidth}}
 \thickhline
 \multirow{2}{*}{\textbf{Model}} & \multicolumn{3}{c}{$\mathrm{NWR}$ }  \\
 \cline{2-4}
 & Simple  & Moderately Complex & Complex \\
 \hline
 \hline
 \emph{Web Search}    & -  & -  & -  \\
 \emph{AI Search} & 0.00\% & 5.00\%$^*$ & \textbf{13.00\%}$^*$  \\
 \thickhline
 \end{tabular}
 }
 \begin{flushleft}
 \footnotesize
 \centering
 $^*$ indicates the statistically significant\\(\emph{t}-test with $p \textless 0.05$ over the legacy system).
\end{flushleft}
\label{online1}
\end{table}
To further investigate the effectiveness of the AI search system in a real-world environment, we perform extensive experiments with real-world traffic and compare it with the legacy system (\ie, web search system) at Baidu Search.

\subsection{Human Evaluation}
To comprehensively evaluate the impact of the proposed AI search system, we conduct a series of human assessments using a side-by-side comparison approach. First, we log a large set of queries and their corresponding search results, which are returned by both the AI search system and the legacy system. Then, we employ professional annotators to judge the superiority of the results from the AI search system compared to those from the legacy system. The manual evaluation results are measured with \emph{Win vs. Tie vs. Lose}, which is a metric measured
by the professional annotators’ judgment. For a query, the annotators are provided with a pair ($\mathrm{Result}_1$, $\mathrm{Result}_2$) whereby one result is returned by system A, and the other is generated by a competitor system B. The annotators, who do not know which system the result is from, are then required to independently rate among \emph{Win} ($\mathrm{Result}_1$ is better), \emph{Lose} ($\mathrm{Result}_2$ is better), and \emph{Tie} (they are equally tied). In order to quantify the human evaluation, we aggregate these three indicators mentioned above as a unified metric, denoted as Normalized Win Rate (NWR):
\begin{align}
    \mathrm{NWR} =\frac{\mathrm{\#Win}-\mathrm{\#Lose}}{\mathrm{\#Win}+\mathrm{\#Tie}+\mathrm{\#Lose}},
\end{align}
where $\#\mathrm{Win}$ (or $\#\mathrm{Lose}$) is the number of results generated from the AI search system better (or worse) than the legacy system, and $\#\mathrm{Tie}$ indicates two results are equally good or bad.
In the manual evaluation, we consider the influence of query complexity on the results for the evaluation. Therefore, we sample three categories of simple, moderately complex, and complex queries to comprehensively compare the two systems. Specifically, the test dataset is constructed in advance by sampling real user queries through two distinct selection strategies as follows:
\begin{itemize}
    \item \emph{Targeted Sampling.} We employ DeepSeek-R1 prompts to assess the complexity of queries and classify them into three categories: simple queries, moderately complex queries, and complex queries.
    \item \emph{Random Sampling.} The test queries are randomly selected, which consist of high-frequency queries, random queries, and long-tail queries.
\end{itemize}

As shown in Table~\ref{online1}, the AI search system delivers substantial improvements to the legacy system, which reveals that the AI search system could better meet user satisfaction. Specifically, for simple queries, the AI search system performs comparably to the legacy system. However, for moderately complex and complex queries, the AI search system demonstrates significant improvements, achieving a relative enhancement of $\mathrm{WTL}$ by 13\% for complex queries. These human evaluation results indicate that the AI search system is capable of delivering accurate and comprehensive responses in complex query scenarios, thereby effectively satisfying user search requirements and validating its effectiveness.

\begin{table}[t]
\centering
\caption{Improvements of the proposed AI search system (\emph{AI Search}) compared with the online legacy system (\emph{Web Search}) on user-side metrics in online A/B Test.}
\setlength\tabcolsep{2pt}
\renewcommand{\arraystretch}{1.3}
 {
 \small
 \begin{tabular}{l | P{0.16\linewidth}| P{0.16\linewidth} | P{0.16\linewidth} | P{0.24\linewidth}}
 \thickhline
 \textbf{Model} & CQR  & PV & DAU & Dwell Time  \\
 \hline
 \hline
 \emph{Web Search} & -       & -       & -     & -\\
 \emph{AI Search} & -1.45\%  & 1.04\%   & 1.85\%   & 0.52\% \\
 \thickhline
 \end{tabular}
 }
\label{online2}
\end{table}

\subsection{Online A/B Test}
To evaluate the real-world effectiveness of the AI search system, we deploy it at Baidu Search and conduct an A/B Test comparing it with the legacy search system. When deploying the AI search system at Baidu Search, specific user-side metrics serve as critical indicators of the effectiveness of search systems in enhancing users' search satisfaction. In the online A/B test, we conduct the experiments with 1.00\% real-world web traffic of Baidu search, focusing on metrics directly related to user engagement and reporting the improvements of the proposed AI search system over the online legacy system. As shown in Table~\ref{online2}, compared to the online legacy system, the AI search system achieves the following relative improvements: the change query rate (CQR) decreases by 1.45\%; the number of page views (PV) increases by 1.04\%; the number of daily active users (DAU) increases by 1.85\%; dwell time (Dwell Time) increases by 0.52\%. All reported values are statistically significant with $p < $0.05. These results suggest that the proposed AI search system substantially enhances user engagement within the search engine context.

\subsection{Case Study}
\begin{figure*}[t]
\centering
\begin{tabular}{@{\ }c@{\ }c}
\includegraphics[width=0.48\textwidth]{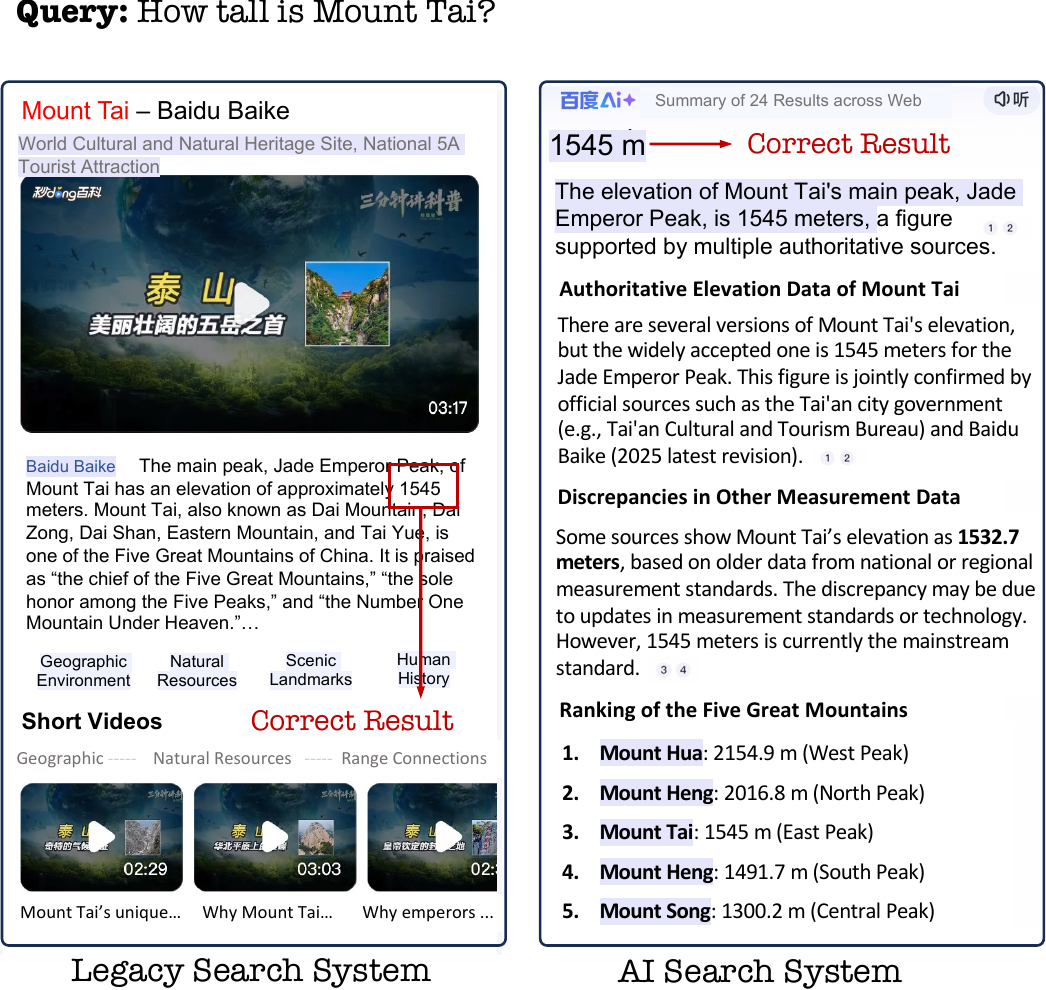} &
\hspace{2pt}
\includegraphics[width=0.48\textwidth]{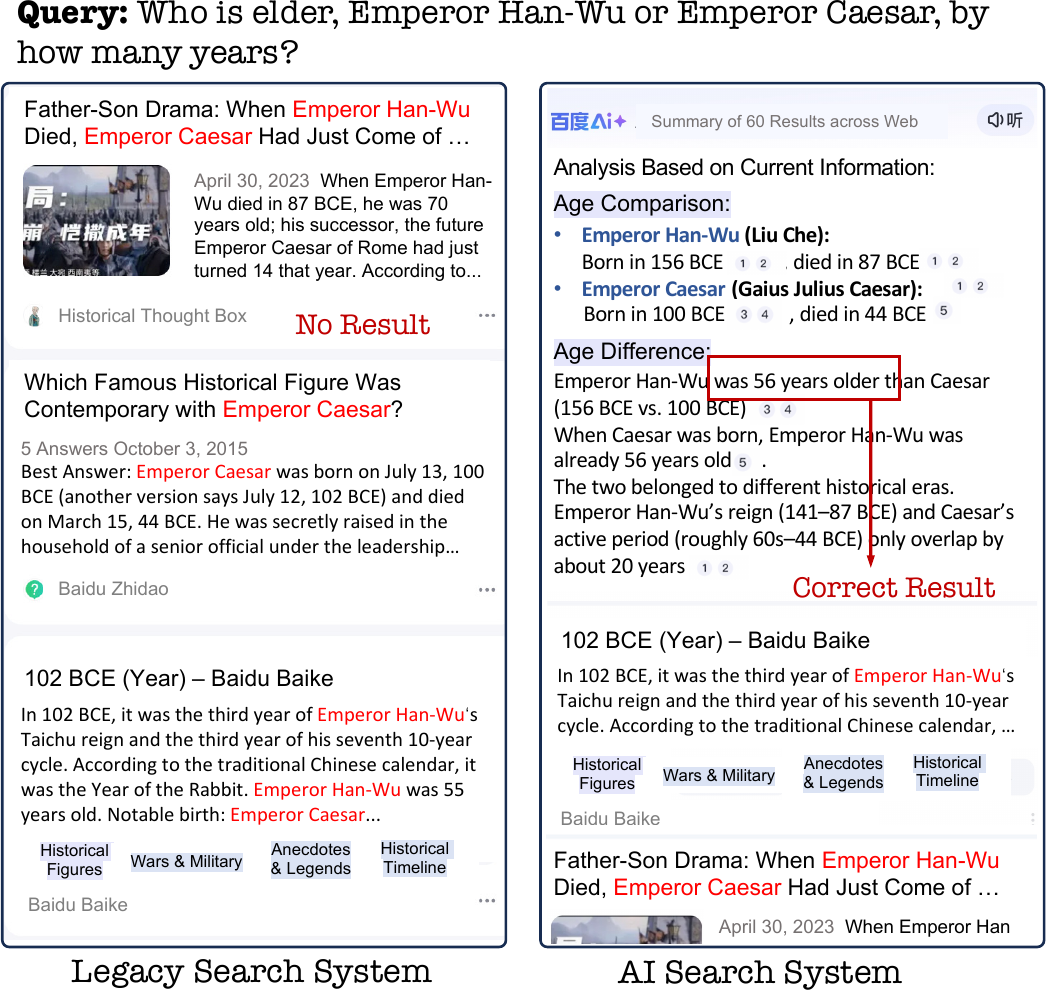} 
\\
{\small (a) A case study of the simple query.} &
{\small (b) A case study of the complex query.}
\end{tabular}
\caption{Online case comparisons of the AI search
system to the legacy search system.} 
\label{case_study}
\end{figure*}

As depicted in Fig.~\ref{case_study}, we present two representative case studies. One case study compares the outcomes of the AI search system with those of the legacy system for a simple query, while the other examines their respective performance for a complex query.
Specifically, for a simple query such as \emph{``How tall is Mount Tai?''}, both the AI search system and the legacy search system promptly provided the correct answer—that ``Mount Tai is 1,545 meters tall''. This case study demonstrates that for straightforward queries, the AI search system can effectively respond by either invoking the web search tool or leveraging its internalized knowledge, while traditional web search systems achieve the correct response through a series of retrieval and ranking processes.
However, for complex queries that require multi-step reasoning, traditional web search methods exhibit notable limitations in fulfilling user requirements and often fail to produce correct answers. For instance, when faced with the complex query ``Who is older, Emperor Han-Wu or Emperor Caesar, and by how many years?'', the legacy search system cannot directly yield the final answer through simple retrieval and ranking procedures. By contrast, the proposed AI search system employs a \emph{Master}-assigned \emph{P}lanner-Enhanced configuration. In this approach, \emph{Planner} first decomposes the main query into sub-tasks and dynamically adjusts the capability boundary by invoking the appropriate tools. Subsequently, \emph{Executor} concurrently carries out two sub-tasks—retrieving the ages of Emperor Han-Wu and Emperor Caesar. Based on the results of these sub-tasks, a calculation tool is then leveraged to compute the age difference, and finally, \emph{Writer} synthesizes the findings to present the final answer. This case study clearly demonstrates the advantages and necessity of the AI search system in effectively addressing complex queries. 
According to these two cases, we find that the AI search system has an enhanced capability to deliver more accurate results for user satisfaction-oriented searches than the legacy system. 

\section{Conclusion}
In this paper, we propose a novel AI search paradigm that fundamentally rethinks the traditional information-seeking process. Our proposed AI search system leverages a modular, multi-agent architecture to emulate and enhance human-like problem solving. Specifically, the Master Agent dynamically assesses query complexity and orchestrates the formation of an agent team, while the Planner Agent decomposes complex queries into a structured DAG of sub-tasks with explicit interdependencies. The Executor Agent then carries out these sub-tasks through the coordinated invocation of diverse tools, and the Writer Agent synthesizes the collected results to generate comprehensive answers.
This multi-agent framework addresses the limitations of classical IR and state-of-the-art RAG systems, which typically provide only a linear, document-retrieval oriented response. By enabling proactive planning, dynamic tool integration, and iterative reasoning, our system can effectively handle complex, multi-step queries, thereby reducing the cognitive load on users and improving overall search quality. The empirical evaluations, as discussed in the experimental sections, further validate the system’s effectiveness in delivering accurate, context-aware responses and enhanced user engagement.
Overall, our work contributes a structured and detailed blueprint that integrates best practices from both industry and academia. It lays the groundwork for future research in AI-driven information seeking and highlights several promising directions for optimizing the performance of collaborative agents and seamless tool integration.

\section*{Acknowledge}
\addcontentsline{toc}{section}{Acknowledge}
We sincerely appreciate all the engineers of Baidu Search for their contributions to the construction of the AI search system.

\bibliography{main}

\end{CJK*}
\end{document}